\documentclass[10pt,sigconf,nonacm]{acmart}
\AtBeginDocument{%
  }
\usepackage{booktabs}
\usepackage{graphicx}
\usepackage{float}
\usepackage{placeins}
\usepackage{graphicx}
\usepackage{multirow}
\usepackage{booktabs}
\usepackage{float}
\usepackage{soul}
\usepackage{makecell}
\usepackage{caption}
\usepackage{enumitem}
\usepackage{pgfplots}
\pgfplotsset{compat=1.18}
\usepackage{pgfplotstable}

\newcommand\benchmarkname{\texttt{\textbf{MM-Telco}}}

\begin{document}

\title{MM-Telco: Benchmarks and Multimodal Large Language Models for Telecom Applications
  }


\author{Anshul Kumar}
\email{anshulchrs@gmail.com}
\orcid{ 0009-0001-4230-8877}
\affiliation{%
  \institution{IIT Bhilai}
  \country{India}
}

\author{Gagan Raj Gupta}
\email{gagan@iitbhilai.ac.in}
\orcid{0000-0002-8568-2949}
\affiliation{%
  \institution{IIT Bhilai}
  \country{India}
}

\author{Manish Rai}
\email{manishr@iitbhilai.ac.in}
\affiliation{%
  \institution{IIT Bhilai}
  \country{India}
}

\author{Apu Chakraborty}
\email{apuchakraborty37@gmail.com}
\orcid{0009-0006-0821-3272}
\affiliation{%
  \institution{IIT Bhilai}
  \country{India}
}

\author{Ashutosh Modi}
\email{ashutoshm@cse.iitk.ac.in}
\affiliation{%
  \institution{IIT Kanpur}
  \country{India}
}

\author{Abdelaali Chaoub}
\email{chaoub.abdelaali@gmail.com}
\affiliation{%
  \institution{INPT}
  \country{Morocco}
}

\author{Soumajit Pramanik}
\email{soumajit@iitbhilai.ac.in}
\affiliation{%
  \institution{IIT Bhilai}
  \country{India}
}

\author{Moyank Giri }
\email{moyankg@iitbhilai.ac.in}
\affiliation{%
  \institution{IIT Bhilai}
  \country{India}
}

\author{Yashwanth Holla }
\email{yashwanthh23@iitk.ac.in}
\affiliation{%
  \institution{IIT Kanpur}
  \country{India}
}

\author{Sunny Kumar }
\email{sunnykumar@iitbhilai.ac.in}
\affiliation{%
  \institution{IIT Bhilai}
  \country{India}
}

\author{M. V. Kiran Sooraj}
\email{mvkiran@iitbhilai.ac.in}
\affiliation{%
  \institution{IIT Bhilai}
  \country{India}
}


\begin{abstract}
Large Language Models (LLMs) have emerged as powerful tools for automating complex reasoning and decision-making tasks. In telecommunications, they hold the potential to transform network optimization, automate troubleshooting, enhance customer support, and ensure regulatory compliance. However, their deployment in telecom is hindered by domain-specific challenges that demand specialized adaptation. To overcome these challenges and to accelerate the adaptation of LLMs for telecom, we propose \benchmarkname, \textbf{a comprehensive suite of multimodal benchmarks and models} tailored for the telecom domain. The benchmark introduces various tasks (both text based and image based) that address various practical real-life use cases such as network operations, network management, improving documentation quality, and retrieval of relevant text and images. Further, we perform baseline experiments with various LLMs and VLMs. The models fine-tuned on our dataset exhibit a significant boost in performance. Our experiments also help analyze the weak areas in the working of current state-of-art multimodal LLMs, thus guiding towards further development and research.

\end{abstract}



\maketitle

\section{Introduction}
LLMs have demonstrated remarkable capabilities across diverse domains, including conversational AI, summarization, and question answering~\cite{Yang2023HarnessingTP}. These models, trained on extensive corpora such as OpenWebText~\cite{openwebtext} and Common Crawl~\cite{commoncrawl}, exhibit strong generalization in open-domain tasks due to their ability to process and generate human-like language, analyze huge amounts of data, and automate advanced decision workflows. Given their generalization capability, LLMs could potentially be useful in the telecom domain for processing telecom-specific documents, logs, and configuration files to assist in troubleshooting, question-answering, and root cause analysis~\cite{shao2024wirelessllmempoweringlargelanguage, 10624781, maatouk2024largelanguagemodelstelecom}. 


However, LLM's direct application to the telecom domain is challenging due to following reasons: \textbf{1) Fine-grained domain adaptation:} Telecom standards and network configurations evolve rapidly, requiring LLMs to accurately process and differentiate between multiple 3rd Generation Partnership Project (3GPP) releases~\cite{3gpp_releases}. General-purpose models often mix knowledge across different versions, leading to inconsistencies in responses~\cite{karapantelakis2024usinglargelanguagemodels}~\cite{bornea2024telcoragnavigatingchallengesretrievalaugmented} \textbf{2) Complex multi-document reasoning:} Many telecom tasks demand cross-referencing multiple sources—such as configuration files, network logs, and standards documentation. Existing LLMs struggle with structured reasoning over these diverse data formats ~\cite{Lior2024SEAM:}~\cite{yu2024unleashingmultihopreasoningpotential}. \textbf{3) Limited multimodal understanding:} Telecom workflows involve a combination of text, network diagrams, and signal visualizations. Most LLMs lack robust image-text reasoning capabilities, making vision-language models (VLMs) an essential but underexplored research area~\cite{Zhang2023Vision-Language}.  \textbf{4) Absence of domain-specific evaluation benchmarks:} Unlike healthcare or finance, telecom lacks well-defined benchmarking frameworks to systematically assess LLM performance on specialized tasks ~\cite{zou2024telecomgptframeworkbuildtelecomspecfic}. Developing structured evaluation datasets is crucial for measuring progress in this domain. \textbf{5) Privacy Concerns:} Proprietary LLMs pose privacy risks, as they require users to share prompts and data with the model's owners~\cite{YAO2024100211}. \textbf{6) Limited Customizations:}  Propriety LLMs offer limited control and customization, as users cannot access model weights. While general-purpose open-source LLMs are adaptable, they often underperform in telecom-specific tasks due to their broad knowledge base, leading to hallucinations and overly generalized outputs \cite{Soman_2023}. 

\textbf{Consequently, a structured evaluation framework is necessary to enable and promote application of LLMs to the telecom domain.} Key areas where LLMs can provide value include multimodal question answering on telecom-related text and image data, automated analysis of network packet captures (PCAPs), and retrieval-augmented generation (RAG) systems for more reliable telecom knowledge retrieval,telecom image generation and updation. Consequently, developing robust \textbf{benchmarks, datasets, and models} is crucial to ensure accuracy, domain-specific adaptation, and effective reasoning over structured data sources.

\begin{figure*}[t]  
    \centering
    \includegraphics[scale=0.09]{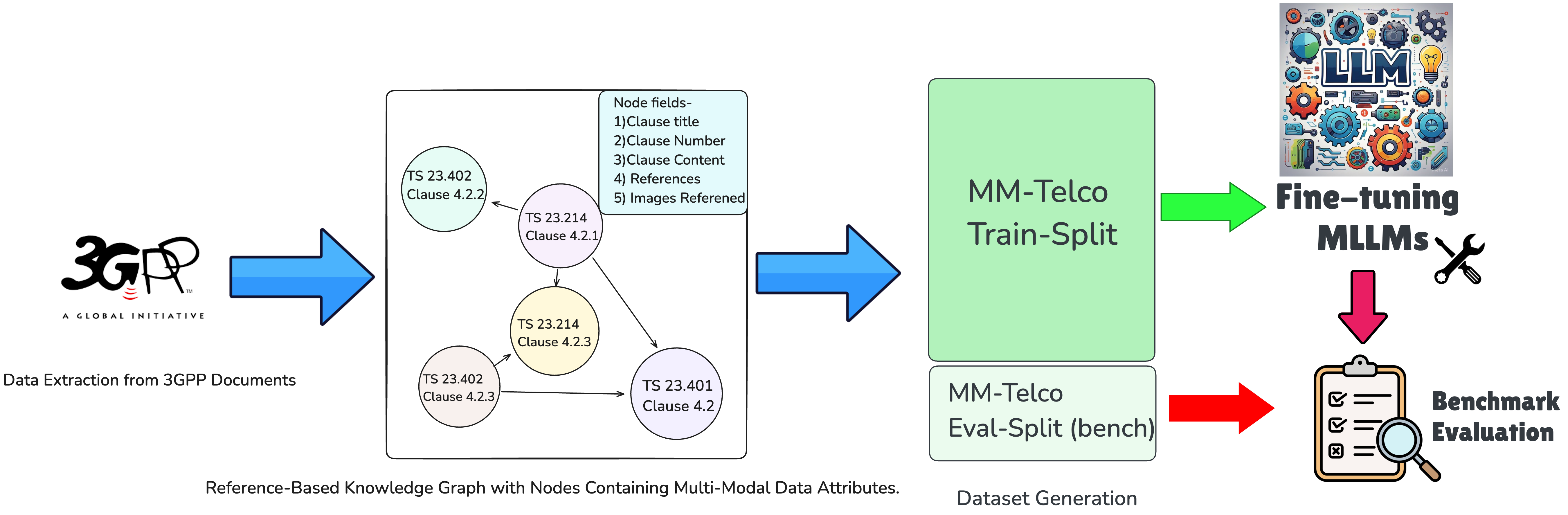} 
    \caption{Key components of our evaluation framework for enabling Multimodal Telecom applications.}
    \label{fig:pipeline}
\end{figure*}

To bridge the above mentioned gaps and enable the adoption of LLMs in telecom, in this paper, we make the following contributions: 

\begin{enumerate}[noitemsep,nosep,leftmargin=*]
    \item \textbf{A multimodal benchmark suite} – We introduce a curated dataset \benchmarkname\ benchmark \footnote{https://github.com/gagan-iitb/MM-TelcoBench/} covering all sections and documents of 3GPP Release 17. It is a structured evaluation framework encompassing both text-based (multiple-choice, long-answer) and image-based (classification, retrieval) tasks, enabling systematic performance assessment of LLMs and VLMs in telecom (details in \S\ref{sec:methodology} and \S\ref{sec:benchmark-overview}). The proposed work overcomes the limitations of existing work in Telecom AI (refer \S\ref{sec:related-work}). 
    \item \textbf{Utility of \benchmarkname:} We motivate the utility of the proposed benchmark via concrete practical research questions inspired from real-life applications in the telecom domain (details in \S\ref{sec:utility}). 
    \item\textbf{A fine-tuned Llama Model} - We introduce a fine-tuned Llama model called \textbf{Llama-VL-Telco} for generating and updating telecom-related images (details in \S\ref{sec:model-training}). This model is able to generate different types of telecom images by giving the right prompt, as illustrated in Figure~\ref{fig:image_gen_sequence}. . 
    \item \textbf{Guidelines for Adapting LLMs} – We Identify the best practices for constructing domain-specific datasets and fine-tuning LLMs to enhance their effectiveness in telecom-related applications (details in \S\ref{sec:model-training}).  
    \item \textbf{Baseline evaluations} – An assessment of existing LLMs and multimodal LLMs (such as GPT-4o, Llama 3.2, and Phi 4) to establish foundational performance benchmarks in telecom AI (details in \S\ref{sec:experiments} and \S\ref{sec:results}). 
\end{enumerate}  
\section{Related Work}\label{sec:related-work}

\begin{table*}[t]
    \centering
    \renewcommand{\arraystretch}{0.8}
    \setlength{\tabcolsep}{3pt}
    \begin{tabular}{p{2.8cm} p{2.8cm} p{3cm} >{\centering\arraybackslash}p{1.5cm} >{\centering\arraybackslash}p{1.5cm} > {\centering\arraybackslash} p{2cm} p{2cm}}
        \toprule
        \textbf{Prior Art} & \textbf{Focus Area} & \textbf{Source} & \textbf{PCAP Analysis} & \textbf{Multimodal} & \begin{tabular}{@{}c@{}}\textbf{Private/} \\ \textbf{Public} \end{tabular} & \textbf{Size} \\
        \midrule
        TeleQnA~\cite{maatouk2023teleqnabenchmarkdatasetassess} & Dataset & 3GPP & \texttimes & \texttimes & Public & 10k QA pairs\\
        TSpec-LLM~\cite{nikbakht2024tspecllmopensourcedatasetllm} & Dataset & 3GPP & \texttimes & \texttimes & Public & 535M words\\
        Cellular-Lint~\cite{298168} & Dataset & 3GPP & \texttimes & \texttimes & Public & -\\
        SPEC5G~\cite{karim-etal-2023-spec5g} & Dataset & 3GPP, blogs & \texttimes & \texttimes & Public & 3.5M Sentences \\
        Tele-Eval~\cite{maatouk2024telellmsseriesspecializedlarge} & Dataset & 3GPP, arXiv, Wikipedia & \texttimes & \texttimes & Public & 750K QA pairs \\
        TeleRoBERTa~\cite{karapantelakis2024usinglargelanguagemodels} & LLM Adaptation & 3GPP & \texttimes & \texttimes & Private & 124M \\
        TelecomGPT~\cite{zou2024telecomgptframeworkbuildtelecomspecfic} & LLM Adaptation & 3GPP, IEEE & \texttimes & \texttimes & Private & - \\
        Telecom Language Understanding~\cite{bariah2023understandingtelecomlanguagelarge} & LLM Adaptation & 3GPP & \texttimes & \texttimes & Private & - \\
        NETBUDDY~\cite{wang2023makingnetworkconfigurationhuman} & LLM Adaptation & - & \texttimes & \texttimes & Private & - \\
        SEEN~\cite{10.1145/3636534.3690678} & ML Assisted Diagnostics & North American Operators & \texttimes & \texttimes & Private & 300K network KPI data \\
        Spotlight~\cite{10.1145/3636534.3649380} & GenAI Adaptation & 5G Open RAN & \texttimes & \texttimes & Private & - \\
        \midrule
        \benchmarkname & Dataset & 3GPP & $\checkmark$ &$\checkmark$&Public&21.5K \\
         \bottomrule      
    \end{tabular}
    \caption{Comparison of Existing Telecom LLMs and Frameworks. While previous work has focused on a single task, we have carefully designed 10 tasks across modalities in \benchmarkname .}
    \label{tab:telecom_llm_comparison} 
\end{table*}

\noindent\textbf{Domain Specific LLMs:} Although general-purpose LLMs demonstrate considerable capabilities in various tasks, their
performance degenerates considerably for tasks which necessitates domain-specific knowledge, such as, solving math problems, problems related to medical, legal, financial and telecom domains. As a result, development of domain-specific LLMs has gained attention in those fields in past few years. LLMs like FinGPT \cite{luukkonen-etal-2023-fingpt}, BloombergGPT \cite{wu2023bloomberggptlargelanguagemodel} in the financial domain, and LawyerGPT~\cite{yao2024lawyer}, InternLM-Law~\cite{fei-etal-2025-internlm} in the legal domain have demonstrated that fine-tuning on curated, domain-specific corpora can significantly enhance model performance. Similar progress in various other domains such as biology~\cite{lam2024largelanguagemodelsplant}, code generation~\cite{roziere2023code}, math~\cite{luo2023wizardmath, satpute2024can, didolkar2024metacognitive}, and healthcare~\cite{Wu2024PMC-LLaMA,yuan2024continued,singhal2023large} further underscores the importance of tailored datasets and domain adaptation strategies. Interestingly, domain adapted LLMs have also opened up new opportunities in the field of mobile task automation\cite{, 10.1145/3636534.3649379}\cite{10.1145/3703323.3704278} where their superior language understanding and reasoning capabilities are allowing users to automate complex and repetitive tasks. 

\noindent\textbf{Multimodal VLMs and Their Relevance to Telecom:}
Modern telecom applications involve structured text, diagrams, tables, and network logs, making multimodal learning critical for improving both interpretability and performance. Research in vision-language models and structured data processing \cite{bai2023qwenvlversatilevisionlanguagemodel,bordes2024introductionvisionlanguagemodeling} indicates that integrating multiple modalities can enhance tasks such as network fault analysis and configuration automation. However, most current telecom evaluations and models remain text-centric, leaving a significant gap in multimodal integration. 
Furthermore, VLMs have shown remarkable success in domains such as healthcare, autonomous driving, and remote sensing \cite{Hartsock2024Vision-Language, Zhou2023VisionLM, Kuckreja2023GeoChat:Grounded}; however, their application in telecom remains unexplored, highlighting a gap in multimodal benchmarking and domain-specific evaluation. 

\noindent\textbf{Existing LLMs and Benchmark Datasets in Telecom:}
Adapting LLMs to the telecom domain involves unique challenges, including the scarcity of open-source pretraining datasets, the prevalence of highly technical documents, and the rapid evolution of telecom standards. As a result, telecom applications have seen relatively limited availability of comprehensive benchmarks and domain specific LLMs. 
While most of the telecom-specific LLM benchmarks are proprietary \cite{zou2024telecomgptframeworkbuildtelecomspecfic,roychowdhury2024understandingdomainadaptedsentence}, open datasets like TeleQnA~\cite{maatouk2023teleqnabenchmarkdatasetassess}, TSpec-LLM~\cite{nikbakht2024tspecllmopensourcedatasetllm}, Cellular-Lint~\cite{298168}, SPEC5G~\cite{karim-etal-2023-spec5g}, Tele-Eval~\cite{maatouk2024telellmsseriesspecializedlarge} and TelBench~\cite{lee2024telbench} offer a foundation for evaluation (see Table.~\ref{tab:telecom_llm_comparison}). However, each of these benchmarks have their individual limitations. For instance, none of them are multi-modal and none of them provides any PCAP analysis data which is highly relevant and useful for telecom domain. Moreover their limited emphasis on 3GPP specifications highlights the need for more diverse and structured benchmarks to assess technical document comprehension in telecom.

In case of LLMs, we observe that general-purpose LLMs such as GPT-3.5 and GPT-4 struggle with intricate, standards-based telecom questions~\cite{maatouk2023teleqnabenchmarkdatasetassess} which necesitates the need for telecom-specific LLMs.
In recent past, Telco-RAG \cite{bornea2024telcoragnavigatingchallengesretrievalaugmented} introduces a retrieval-augmented generation framework that enhances accuracy by optimizing query enhancement and document retrieval from technical sources. TelecomGPT \cite{zou2024telecomgptframeworkbuildtelecomspecfic} further extends these efforts by providing a comprehensive framework that includes continual pre-training, instruction tuning, and new evaluation benchmarks, thereby demonstrating superior performance in telecom-specific tasks.
Similarly, 5G Instruct Forge~\cite{10794684} presents a data engineering pipeline that converts complex 3GPP specifications into training datasets, thereby improving model performance on 5G tasks. 
In~\cite{lin2023pushing},  LLMs are effectively deployed in end user locations and vicinity thorough edge training and edge inference.
Furthermore, Tele-LLMs \cite{maatouk2024telellmsseriesspecializedlarge} address the gap by creating a series of specialized models leveraging the Tele-Data and Tele-Eval datasets to capture telecom-specific terminology and mathematical representations, ultimately outperforming general-purpose models while mitigating catastrophic forgetting.
Despite these advances, challenges remain in integrating multimodal data and standardizing benchmarks across diverse telecom applications, underscoring the need for continued research in domain adaptation.

\section{Methodology}
\label{sec:methodology}

Recent advances in LLMs have demonstrated strong generalization capabilities, yet their application in telecom remains underexplored due to the lack of domain-specific benchmarks. To address this gap, we introduce \textbf{MM-Telco}, a multimodal telecom benchmark designed to evaluate LLMs and VLMs across key telecom-specific tasks. Our benchmark systematically assesses models on structured question answering, retrieval, and reasoning over telecom data, including text-based 3GPP documents, network logs, and images. 

The benchmark creation follows a structured pipeline (see Fig.~\ref{fig:pipeline}), consisting of the following stages:

\begin{enumerate}[noitemsep,nosep,leftmargin=*]
    \item \textbf{Structured Data Extraction:} We extract information from 3GPP technical specification (TS) documents, which serve as a key knowledge source for telecom protocols. These documents are parsed to create a structured representation of subclauses, capturing attributes such as subclause title, number, content, references, and images.

    \item \textbf{Knowledge Graph Construction:} The extracted subclauses are organized into a knowledge graph where nodes represent subclauses and edges indicate referenced documents in the subclause, enabling structured reasoning over telecom regulations and specifications.

    \item \textbf{Dataset Creation:} Using the knowledge graph, we generate \textbf{MM-Telco} benchmark, a benchmark dataset to systematically evaluate the performance of fine-tuned models. The training dataset \textbf{MM-Telco Train} is derived from it. 
    
    \item \textbf{Model Fine-tuning and Evaluation:} Pretrained VLMs are fine-tuned using MM-Telco Train, resulting in domain-adapted models. The fine-tuned models are then evaluated on MM-Telco benchmark to assess their reasoning and retrieval performance over telecom data.
\end{enumerate}

This methodology enables the development of robust, telecom-specialized VLMs, ensuring effective domain adaptation while minimizing computational overhead.

\subsection{MM-Telco Benchmark}
\label{sec:benchmark-overview}
Benchmarks enable meaningful comparisons between various LLM models as well as between LLMs and telecom experts, driving advancements in the field by identifying strengths and limitations. MM-Telco encompasses a diverse set of tasks designed to test the capabilities of LLMs and VLMs in the telecom domain, facilitating progress through systematic evaluation and model improvement. Table~\ref{tab:mm_telcobench} summarizes the various tasks, dataset sizes, and the corresponding cognitive skills being evaluated. The benchmark includes both textual and multimodal tasks, ensuring comprehensive evaluation of reasoning, knowledge understanding, and retrieval capabilities.

\begin{table*}[t]
    \centering
    \renewcommand{\arraystretch}{1.2}
    \setlength{\tabcolsep}{2pt}
    \begin{tabular}{llll}
       \toprule
        \textbf{Category} & \textbf{Task Type} & \textbf{Dataset Size} & \textbf{Skill Tested} \\
        \midrule
        \multirow{4}{*}{\hyperref[fig:text_example]{Text}}
        & Multi-Choice QA  & 10,000 Qs & Knowledge Recall and Understanding \\
        & Multi-Hop Multi-Choice QA  & 2,000 questions &   Knowledge Recall and Understanding \\
        & Long-Form Telecom Blog QA  & 1,000 Qs & Comprehension \\
        & Long-Form QA  & 2,500 Qs & Comprehension \\
        & Information Retrieval  & 
        \begin{tabular}{@{}c@{}}1,000 queries, \\ 10,000 docs\end{tabular}
         & Search and Reasoning \\
        & Named Entity Recognition (NER) & 1,000 entities & Entity Identification \\
         & Scenario Based Filter Generation & 500 Qs & Wireshark filter suggestion for PCAP diagnosis\\
        \cmidrule(lr){1-4}
        \multirow{4}{*}{\hyperref[fig:image_example]{Image}}
        & Image-based Multi-Choice QA  & 2,000 Qs &  Visual Understanding \\
        & Image Caption Generation & 1,000 images &  Image-Language Mapping \\
        & Image-Based Long QA & 1,000 Qs  & Multimodal Reasoning \\
        & Image Retrieval & 1,000 queries, 3,000 images &  Content-Based Search \\
        & Image Generation & 7,000 mermaid code and images  & Image Generation\\
        & Image Editing & 11,000 images + prompt& Image understanding and Editing\\
        \bottomrule
    \end{tabular}
    \caption{Summary of all the tasks in MM-Telco}
    \label{tab:mm_telcobench}
\end{table*}

\subsection{Telecom Image Generation and Editing}
The  challenge of accurately interpreting and correcting discrepancies in protocol diagrams commonly found in telecom standards. Ensuring precise and consistent representations of these diagrams is crucial for telecom engineers to effectively understand, implement, and adhere to protocol specifications, thus minimizing miscommunication and errors in network deployments.
So,to address this challenge,we fine-tune a Llama model \textbf{Llama-VL-Telco} that is capable of generating and updating the telecom images when given the suitable prompt.This model was trained on the curated dataset that consisted of pair of mermaid codes and their corresponding images.For training of Image generation task,we parse Mermaid.js code and generate a detailed topological description for each diagram type using a templated approach, ensuring accuracy and minimizing hallucination errors. Once the model learns code generation from images, we advance further by enabling the model to generate code directly from these topological descriptions.

For training of Image correction task,we removed some of the nodes or edges from the mermaid code which resulted in the incomplete image and then this incomplete image along with an improvement prompt was treated as query while the original image was treated as the ground truth.

\subsection{Processing 3GPP Documents}
\label{sec:structured-data-extraction}

\textbf{3GPP Technical Specification (TS) Documents} serve as the foundational standard for modern telecom networks, defining protocols, architectures, and signaling procedures for technologies such as LTE, 5G, and beyond. These documents are structured hierarchically into numbered clauses and subclauses, with extensive inter-references across specifications. Standard dataset construction methods often apply naive chunking approaches to these documents, leading to loss of semantic continuity and incomplete information extraction. 

To overcome these limitations, we develop a structured extraction framework that preserves contextual integrity by leveraging the hierarchical structure of 3GPP documents. Our pipeline extracts subclauses along with their referenced content (textual clauses and figures) and represents them in a structured format. This approach ensures that generated question-answer pairs maintain semantic coherence and retain crucial cross-references.  

References within the subclauses of a 3GPP document are categorized into four types:

\begin{enumerate}
    \item \textbf{Local References:} Citations within the same document (e.g., Clause 4.3, Subclause 3.5).  
    \item \textbf{Global References:} Citations to other 3GPP documents (e.g., Clause 5.2 in TS 39.402).  
    \item \textbf{Document References:} References to entire 3GPP specifications without specific clause details (e.g., 3GPP TS 23.501).  
    \item \textbf{Image References:} Embedded diagrams and figures (e.g., Figure 7.5.1-1) essential for network architecture explanations.  
\end{enumerate}  

This structured data extraction process serves as the foundation for generating diverse tasks in MM-Telco, enabling precise benchmarking of LLMs in telecom-specific applications. Figure \ref{fig:text_example} and Figure \ref{fig:image_example} present examples of text-based and image-based questions, respectively.
 
\subsubsection{Multiple Choice Questions}
\textbf{Motivation} - Multiple Choice Questions (MCQs) help evaluate the factual knowledge of LLMs. Additionally, generating MCQs based on different Technical Specification (TS) documents within the same working group enables assessment of the LLMs' performance across various domains in telecom, as shown in Table \ref{tab:3gpp_wgs}.

\begin{figure}[h]
    \centering
    \includegraphics[width=0.2\textwidth]{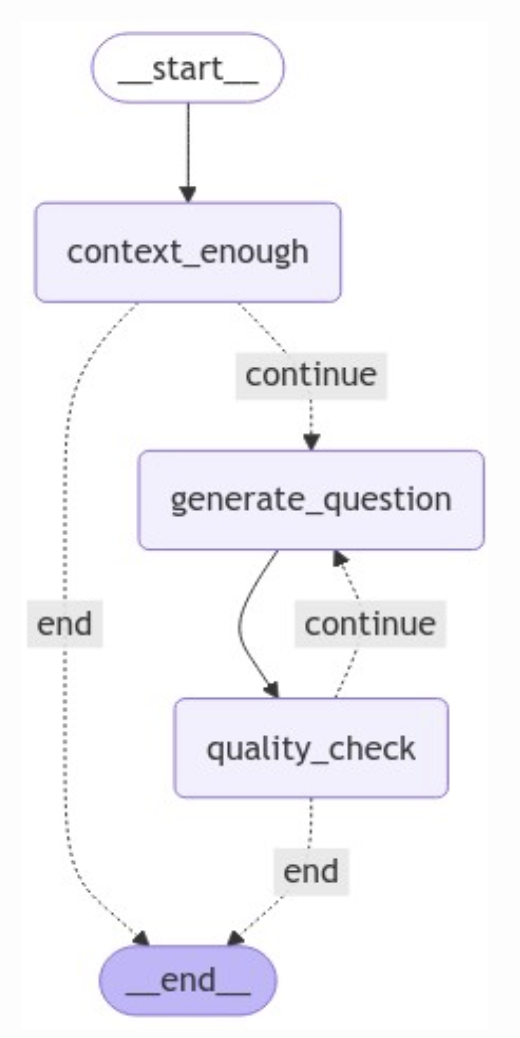}
    \caption{Agentic Pipeline for MCQs generation}
    \label{fig:agent}
\end{figure}

\noindent\textbf{Task Creation Process} - We generated MCQs using structured data extracted from clauses and subclauses of 3GPP Technical Specification documents. The process utilized Nemotron 70B from Nvidia, interfaced through Ollama, within an agentic pipeline. This pipeline consists of three agents: 1) The first agent evaluates whether the given context is suitable for generating high-quality MCQs. If the context lacks sufficient information or contains excessive references, it is discarded. This step ensures that only meaningful clauses are considered. 2) The second agent generates the MCQ based on the selected context. 3) The third agent, the evaluator (gatekeeper), assesses whether the generated MCQ meets predefined quality criteria. If the question fails, the evaluator provides feedback to the generator agent for refinement. This iterative process continues for a maximum of three attempts; if the question still does not meet the criteria, the context is discarded. This architecture ensures the generation of high-quality MCQs through iterative refinement. Figure \ref{fig:agent} visualizes the overall pipeline.

\noindent\textbf{Skills Tested} - The MCQs assess factual correctness and the ability to reason across different 3GPP working groups. They evaluate competencies in various domains such as protocols, service management, security, and privacy.

\begin{figure}[h]
    \centering
    \includegraphics[scale=0.15]{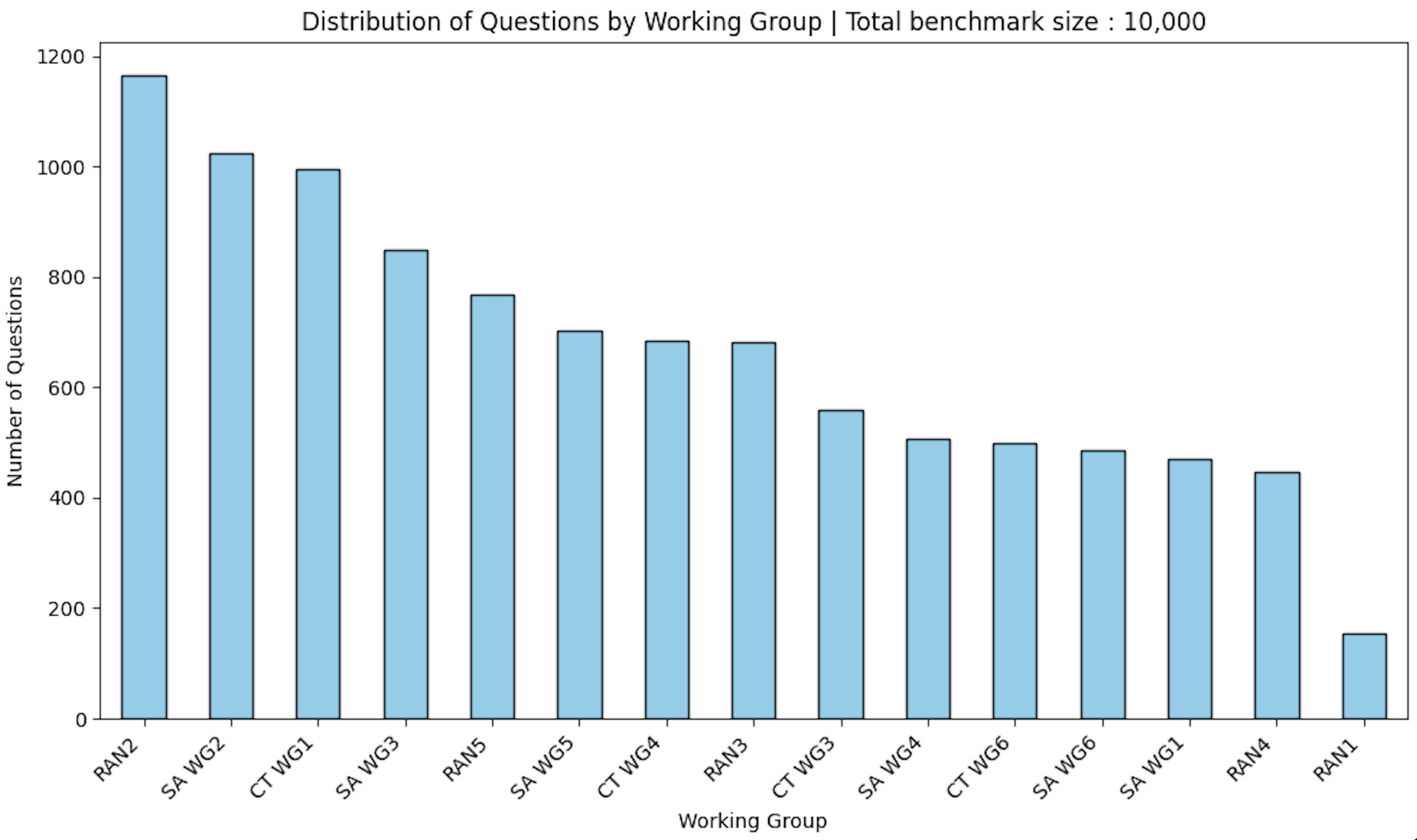}
    \caption{Distribution of MCQ questions across 3GPP working groups}
    \label{fig:10k_dis}
\end{figure}

\subsubsection{Multihop MCQ Questions}

\noindent\textbf{Motivation} - A capable LLM should not only be able to answer MCQs based on a single subclause of one document but also handle questions that require reasoning over linked clauses from multiple TS documents. The need for such a task arises due to fragmented context, inconsistent knowledge retention, and the necessity for precise logical inference. Evaluating performance on multihop MCQs provides insights into an LLM’s ability to process complex, structured telecom knowledge, which is crucial for automation and troubleshooting.

\noindent\textbf{Task Creation Process} - We utilized subclauses with global references and incorporated them as context to generate multihop MCQs. The architecture follows the same agentic pipeline as standard MCQs, with one key difference: we employed the QwQ reasoning model from Alibaba AI. This model was tasked with identifying bridge entities between documents and generating questions based on this interconnected information.

\noindent\textbf{Skills Tested} - Multihop MCQs in our benchmark assess an LLM's ability to integrate information from multiple telecom documents. Given the distributed nature of 3GPP technical specifications, this skill is essential. Unlike standard MCQs, multihop questions require reasoning across two or more references.

\noindent\textbf{Data Statistics} - We generated a total of 2000 multihop MCQs.

\subsubsection{Long-Form QA}

\noindent\textbf{Motivation} - Long-answer questions test an LLM's ability to understand, explain, and connect complex information. Unlike MCQs, they require detailed answers that combine ideas from different parts of the content. This is important for tasks such as writing technical documents, providing troubleshooting guidance, and explaining telecom concepts clearly.

\noindent\textbf{Task Creation Process} - To generate comprehensive long-answer questions, we utilized structured data from 3GPP documents. We sampled clauses along with their referenced content, including both local and global subclauses. We employed GPT-4o from OpenAI to generate question-answer pairs in two steps. First, GPT-4o was instructed to identify all relevant information from up to eight subclauses and create a blueprint outlining how the question should be framed and what the answer should contain. In the second step, GPT-4o was guided to generate the final QA pair using this blueprint. This method ensures high-quality questions by leveraging a broader set of subclauses for more comprehensive answers. Additionally, the blueprint serves as a set of guidelines for evaluating responses. These \textbf{guidelines can be used as rubrics} when instructing another LLM to score the answers, following the \textbf{LLM-as-a-judge} approach. This is particularly valuable since traditional metrics like ROUGE scores are often insufficient for evaluating long-answer questions \cite{xu2023criticalevaluationevaluationslongform}, making such an improved evaluation method crucial. \textbf{GPT-4o-mini} is used as the evaluation judge, instructed to assign a score between 0 and 100 to the generated answer based on the provided rubrics. We also utilized SEM score, which is the cosine similarity between the ground truth answer and the generated answer, using the embedding model BAAI/bge-small-en-v1.5.

\noindent\textbf{Skills Tested} - This task assesses the model's ability to understand technical details, establish logical connections, and provide clear and accurate explanations on telecom topics.

\noindent\textbf{Data Statistics} - We generated a total of 1,500 long-answer question-answer pairs.

\subsubsection{Long-Form Telecom Blog QA}

\noindent\textbf{Motivation} - Long-answer questions assess an LLM's ability to understand, explain, and connect complex information. While generating QA pairs using 3GPP clauses and subclauses ensures high-quality, specification-specific questions, evaluating broader telecom knowledge, evaluating such as an overview of 5G requires a different approach. To assess the model's understanding at a higher conceptual level, we leveraged popular telecom blogs to generate relevant question-answer pairs.

\noindent\textbf{Task Creation Process} - We scraped content from well-known telecom blogs such as Telecom Hall and ShareTechnote. Using Microsoft's Phi-4 model, we provided the entire blog content as context and instructed the model to generate as many relevant questions as possible. In the next step, we prompted the model to generate answers for those questions using the provided blog content.

\noindent\textbf{Skills Tested} - This task evaluates the model's ability to comprehend technical content, synthesize information from broader telecom topics, and deliver clear, well-structured explanations.

\noindent\textbf{Data Statistics} - We generated a total of 1,000 long-answer question-answer pairs.

\subsubsection{Information Retrieval}

\noindent\textbf{Motivation} - Retrieval Augmented Generation (RAG) is an effective method for improving LLM responses by assisting LLMs with relevant documents/content for a given query to reduce errors in answer generation. This helps to reduce hallucination and is one of the cheapest and most straightforward ways to incorporate domain-specific/user-specific knowledge in LLM question answering without needing any complicated expensive process like fine-tuning LLMs. In RAG, the user documents are first embedded using an embedding model into vector representations. When a user asks a query, the most cosine-similar document relevant to the query is retrieved and provided to the LLM, supplementing the final answer generated. Choosing the right embedding model and its performance in the telecom domain is very crucial for effective implementation of RAG for telecom.

\noindent\textbf{Task Creation Process} - We generated questions on various clauses of the technical specification documents, for question generation we used GPT-4o. Now the questions act as the queries, the documents on which the questions were generated act as the documents and the mapping is the qrels, i.e., the ground truth. We included additional documents that did not have any questions generated on them to act as noise. Evaluation is done by calculating the percentage of queries where the relevant document appears within the top K retrieved text results.

\noindent\textbf{Skills Tested} - The ability of the embedding model to generate good vector representations of telecom content. These representations should be distinguishable from embeddings created for documents of  different domains within 3GPP.

\noindent\textbf{Data Statistics} - The dataset includes a total of 1000 queries and qrels and 10,000 documents.

\subsubsection{Named Entity Type Classification}

\noindent\textbf{Motivation} - Accurate identification and classification of telecom-related entities are essential for improving data understanding and enhancing information processing. Within the 3GPP domain, technical content often references various entity types that must be categorized to ensure clarity and consistency in documentation, analysis, and model performance evaluation.

\noindent\textbf{Task Creation Process} - We first utilized the extracted clauses and subclauses from 3GPP documents to identify key telecom entities. After analyzing these entities, we collaborated with subject matter experts to define a comprehensive set of telecom-specific entity types. These categories were designed to cover various technical aspects relevant to 3GPP specifications. The defined entity types include: ['Message', 'Parameter', 'Interface', 'Connection', 'Network Function', 'Protocol', 'Measurement', 'Procedure', 'Specification', 'Technology', 'Network', 'Device', 'Service', 'Organization', 'Working Group', 'Protocol', 'State', 'Item', and 'Mode'].

\noindent\textbf{Skills Tested} - This task evaluates the model’s ability to correctly identify and classify technical terms across multiple telecom-related domains, ensuring a precise understanding of 3GPP terminology and concepts.

\noindent\textbf{Data Statistics} - The dataset includes a total of 1,000 named entities spanning all defined categories to ensure a comprehensive evaluation of the model's classification capabilities.

\subsubsection{Scenario Based Filter Generation}

\textbf{Motivation} PCAP (Packet Capture) analysis plays a vital role in telecom troubleshooting by providing detailed insights into network traffic. Different telecom protocols require specific troubleshooting steps, which often involve applying appropriate Wireshark filters to isolate issues effectively. Creating scenario-based filters enhances the ability to diagnose and resolve telecom specific network problems efficiently.

\textbf{Task Creation Process} We scraped telecom protocol specific Wireshark filters from the official Wireshark website. Using GPT-4o, we generated various scenarios where these protocol-specific filters could be applied for diagnosis. Next, we instructed GPT-4o to create question answer pairs based on PCAP analysis scenarios, suggesting relevant troubleshooting steps and appropriate filters for effective diagnosis.

\textbf{Skills Tested} This task evaluates the model's ability to recommend accurate troubleshooting steps while selecting the appropriate filters for diagnosing telecom scenarios through PCAPs.

\textbf{Data Statistics} The dataset has 500 scenario based question answer pairs.

\subsubsection{Image-Based Multiple Choice Questions (MCQ)}

\noindent\textbf{Motivation} - Image based MCQs test a model's ability to understand visual information related to telecom concepts in 3GPP images. By analyzing diagrams, flowcharts, and system architectures, the model must extract key insights to answer questions accurately. This method effectively evaluates a model's comprehension of telecom-specific visual content.

\noindent\textbf{Task Creation Process} - We generated detailed descriptions of telecom-related images found in 3GPP documents as shown in Figure \ref{fig:pcap_example}. These descriptions consolidate key information from the images and their referenced content to ensure comprehensive coverage. Using these descriptions, we created MCQs that target crucial telecom concepts such as network procedures, system components, and performance metrics. For instance, flowchart-based questions may emphasize decision-making steps or state transitions, while architecture diagram-based questions may focus on system components and their interactions.

\noindent\textbf{Skills Tested} - The task assesses the model's ability to interpret visual content, understand system flows, and analyze network interactions. It also evaluates the model’s understanding of technical concepts presented visually.

\noindent\textbf{Data Statistics} - The dataset includes a variety of image-based MCQs, having total of size 2000.

\subsubsection{Image-Based Long Answer Questions}

\noindent\textbf{Motivation} - Image-based long-answer questions evaluate a model's ability to provide detailed explanations for telecom related visual content. This is crucial for assessing the model’s capacity to understand complex system flows, component roles, and interactions depicted in diagrams. Also generate long details explanations for them.

\noindent\textbf{Task Creation Process} - Firstly we have generated the descriptions for the images by passing the image, context and caption to the Janus-Pro-7B model and instructed it to generate detailed structured descriptions. We utilized detailed descriptions of telecom related images from 3GPP documents to generate long-answer questions using phi-4 model. We passed the descriptions to phi-4 model and instructed it to generate concise and detailed questions, and then passed these questions to generate accurate answers. These questions were designed to reflect real-world scenarios where users may require comprehensive explanations. The questions extend beyond structural identification, focusing on the significance of system flows, component roles, and intricate interactions shown in the diagrams.

\noindent\textbf{Skills Tested} - The task evaluates the model’s ability to analyze visual content, explain technical concepts in depth, and provide coherent, well-\ structured responses to complex telecom scenarios.

\noindent\textbf{Data Statistics} - The dataset includes a variety of image-based long-answer questions, ensuring diverse coverage of telecom concepts and real-world problem-solving scenarios. Total size of the dataset is 2000.

\subsubsection{Image Retrieval}

\noindent\textbf{Motivation} - Accurate image retrieval in the telecom domain is crucial for enhancing understanding and improving accessibility to complex technical content. With the diverse range of image types in 3GPP documentation, effective retrieval requires multimodal embedding models that integrate both visual and textual information. Unlike traditional keyword-based search, multimodal embeddings bridge the gap between text queries and relevant images, enabling precise and efficient information retrieval. Models like ColPali\cite{faysse2025colpaliefficientdocumentretrieval} and CLIP (Contrastive Language Image Pretraining)\cite{radford2021learningtransferablevisualmodels} demonstrate this capability by mapping text and image data into a common vector space, improving search accuracy.

\noindent\textbf{Task Creation Process} -We constructed our benchmark by pairing structured queries with corresponding images from 3GPP documents. These queries were carefully designed using GPT-4O model to reflect the contextual significance of telecom diagrams, including network procedures, system components, and performance metrics. To enhance the challenge, we incorporated additional images unrelated to the queries, creating a realistic retrieval scenario that demands precise matching based on technical context.

\noindent\textbf{Skills Tested} - This task evaluates the ability of multimodal embedding models to generate meaningful vector representations for both textual descriptions and telecom related images. Effective embeddings should capture technical details, ensuring accurate alignment between  the visual content and its corresponding text based query. Evaluation is done by calculating the percentage of queries where the correct image appears within the top K retrieved images.

\noindent\textbf{Data Statistics} - The dataset includes a total of 50 structured queries and 3766 images, ensuring comprehensive coverage of diverse telecom concepts and visual representations.

\subsubsection{Image Caption Generation}

\noindent\textbf{Motivation} Image caption generation is a crucial task in computer vision that involves generating concise yet informative descriptions of visual content. In the telecom domain, accurately summarizing technical diagrams is essential for improving document accessibility and enhancing understanding. Visual Language Models (VLMs) play a key role in this by producing captions that highlight critical components, system interactions, and technical details present in telecom related images.

\noindent\textbf{Task Creation Process} To build this benchmark, we extracted the image descriptions and captions provided directly under the images in 3GPP technical specification documents. These extracted captions were used as ground truth references to evaluate the model's ability to generate accurate and meaningful descriptions that convey key telecom related details.

\noindent\textbf{Skills Tested} This task evaluates the model's ability to understand and describe technical diagrams effectively. Successful models should demonstrate an understanding of technical content by generating captions that highlight essential elements such as network procedures, system architecture, or component interactions.

\noindent\textbf{Data Statistics} The dataset includes 1000 images paired with corresponding extracted captions, providing a focused evaluation of telecom specific visual content.

\section{Model Training } \label{sec:model-training}

Our approach to developing telecom-specific multimodal LLM consists of three key steps: training dataset creation, various downstream tasks, and efficient training techniques. Each of these steps play a crucial role in ensuring that the model can effectively process and generate telecom-related information while maintaining general reasoning capabilities.

One of the primary challenges in building a domain-specific LLM is the creation of a high-quality telecom-related dataset. The dataset must encompass technical documents, network configurations, troubleshooting logs, and multimodal data, ensuring comprehensive coverage of telecom concepts. The second major challenge is mitigating catastrophic forgetting, where the model, when fine-tuned on telecom data, risks losing general knowledge. Careful dataset selection and continual learning techniques are essential to balance domain adaptation with broad knowledge retention. The third challenge is the significant computational resources required to fine-tune a multimodal LLM effectively. Training large models with textual and visual data demands high-memory GPUs, optimized training pipelines, and parameter-efficient fine-tuning techniques to make the process feasible while maintaining model performance and efficiency.

In the following sections, we will provide details of each step.

\subsection{Experimental Setup} \label{sec:experiments} 
All experiments were conducted on an NVIDIA RTX A6000 GPU using PyTorch with CUDA 11.2 and cuDNN 8.1 under a Linux environment. We fine-tuned Llama 3.1 8B Instruct  and Qwen 2.5 vl 7B Instruct via Low-Rank Adaptation (LoRA), applying rank R=256 and alpha = 512. Training employed batch size of 1 and learning rate of 0.01 using FP16 mixed precision. We have evaluate the models on different benchmarks to evaluate the model capability on telecom domain. 

\noindent\textbf{Training Approach}
To develop the  telecom specific models, we utilized LLaMA 3.1 8B Instruct as the base model for text specific task and Qwen 2.5 vl 7B Instruct as the base model for Image specific task. To adapt the model to the telecommunications domain while mitigating high computational costs, we employed parameter-efficient fine-tuning techniques. Specifically, we implemented Low-Rank Adaptation (LoRA), which enables efficient adaptation by injecting low-rank updates into pre-trained model weights. This approach significantly reduces the number of trainable parameters while maintaining performance, making fine-tuning more feasible for large-scale telecom applications.

\noindent\textbf{Parameter Efficient Fine Tune (PEFT):}
To address the computational challenges associated with fine-tuning a multimodal LLM, we have implemented the Low-Rank Adaptation (LoRA) method as part of Parameter-Efficient Fine-Tuning (PEFT)
\cite{hu2021loralowrankadaptationlarge}. Traditional fine-tuning requires updating all model parameters, which is highly resource-intensive, especially for large-scale multimodal models. LoRA significantly reduces computational costs by introducing trainable low-rank matrices into the model’s existing weights, allowing adaptation with minimal parameter updates.
In the LoRA method, it is important to be aware of the rank and alpha values, as choosing optimal values is crucial for achieving a better domain-specific model. In our case, we have set the rank to 256 and the alpha value to 512.

\section{Utility of the Proposed Benchmark for Telecom Industry}\label{sec:utility}

Our domain-specific benchmarks address critical research questions that directly affect telecom operations in real-world settings. These use cases not only validate our experimental framework but also offer early adopters actionable insights into deploying advanced LLMs and VLMs in operational networks\\~\\
\textbf{RQ1. Embedding Models for RAG Systems:} \\
\textbf{Research Question:} Which embedding models deliver robust performance in retrieval-augmented generation (RAG) systems for telecom applications? \\
\textbf{Importance:} Accurate retrieval across heterogeneous telecom data—spanning both text and images—is critical for efficient information access and decision-making. \\
\textbf{Experimental Outcomes:} Our experiments evaluated multiple embedding models using top-1, top-3, and top-5 accuracy metrics, pinpointing the optimal configuration that significantly enhances retrieval precision in a multi-modal RAG system.\\~\\
\textbf{RQ2. Autonomous Network Management and PCAP Analysis Tool:} \\
\textbf{Research Question:} How can AI agents automate complex network management tasks such as troubleshooting and incident resolution using PCAP Analysis to maintain optimal performance? \\
\textbf{Importance:} Automation of network management tasks not only reduces human error but also dramatically cuts down incident response times. By leveraging AI trained on rich, telecom-specific data, operators can resolve issues faster and ensure uninterrupted service—a real-world advantage that translates into enhanced customer satisfaction and lower operational costs. \\
\textbf{Experimental Outcomes:} We show that LLMs, trained on 3GPP text and images, effectively address troubleshooting queries typically handled by network engineers. Our RAG-based ChatBot model, which supports both textual and visual inputs, demonstrates a marked improvement in incident resolution speed and accuracy.

\noindent\textbf{RQ3: Private Fine-Tuned LLMs for Telecom}\\
\textbf{Research Question:} How does a locally fine-tuned telecom-specific LLM enhance performance while ensuring data privacy compared to general-purpose models?\\
\textbf{Experimental Outcomes:} We present a fine-tuned local telecom-LLM trained on domain-specific textual data, offering improved performance in telecom-related tasks while preserving data privacy. By operating within a controlled local environment, the model mitigates risks associated with transmitting sensitive telecom data to external servers. Experimental results indicate superior accuracy in technical document comprehension, troubleshooting, and structured knowledge retrieval, demonstrating the advantages of localized fine-tuning for both performance and security.  

\noindent\textbf{RQ4. Text-to-Image Generation for Documentation:} \\
\textbf{Research Question:} How can AI automatically generate precise images for technical documentation, troubleshooting guides, and other telecom-related visual content? \\
\textbf{Importance:} High-quality, consistent visual representation in documentation is essential for effective communication among network engineers and field technicians. Automating the generation of technical diagrams reduces manual effort, minimizes errors, and speeds up the documentation process. \\
\textbf{Experimental Outcomes:} We demonstrate that the model can fix inconsistencies and regenerate accurate  different types telecom diagrams images using Mermaid code and suitable prompts.

\noindent\textbf{RQ5. Real-Time AI Agent Support:}\\
\textbf{Research Question:} How can AI support real-time customer interactions and reduce manual efforts in routing and responding to communications? \\
\textbf{Importance:} In today’s fast-paced telecom landscape, every minute of delayed customer support can translate into significant revenue loss and customer dissatisfaction. AI-driven, real-time chatbots can process and respond to customer queries instantly, ensuring continuous support without overburdening human agents. This improves service quality and operational efficiency, making a tangible impact in customer-facing scenarios. \\
\textbf{Experimental Outcomes:} Our RAG-based ChatBot model demonstrates continuous, real-time support by efficiently handling customer inquiries and routing communications, thereby significantly reducing manual intervention and improving overall response times.\\~\\
Overall, these use cases validate the transformative potential of our domain-specific benchmarks. They not only address pressing research questions but also offer compelling, real-world benefits that early adopters in the telecom sector can harness to achieve enhanced operational efficiency, cost savings, and superior customer service.
\section{Results}\label{sec:results}
In this section, we analyze and discuss the evaluations of various LLM models on our proposed benchmark and address the questions raised in Section \ref{sec:utility}.

\noindent\textbf{RQ1: Embedding Models for RAG Systems:}
Our experiments evaluated several embedding models to determine which deliver the best performance for retrieval-augmented generation (RAG) in telecom applications. As shown in Table~\ref{tab:mm_telcobench_train}, models such as \textbf{bge-large-en-v1.5} and \textbf{mxbai-embed-large-v1} achieved the highest top-1, top-3, and top-5 accuracy scores, indicating their superior capability in retrieving relevant telecom information. In contrast, models like \textbf{snowflake-arctic-embed-I} demonstrated significantly lower performance, highlighting the importance of selecting embedding models that are well-suited for the complexities of telecom data. These results underscore the need for embedding models optimized for domain-specific tasks, ensuring robust retrieval in multi-modal RAG systems.

\noindent\textbf{RQ2: Autonomous Network Management and PCAP Analysis Tool:} 
\noindent\textbf{RQ2: Autonomous Network Management and PCAP Analysis Tool:} 
In this section, we evaluate the capability of AI models in automating complex network management tasks, including troubleshooting, incident resolution, and PCAP-based analysis. Packet capture (PCAP) files contain detailed network traffic data, requiring AI models to interpret and generate appropriate filtering rules for anomaly detection and security enforcement. 

Table~\ref{tab:mm_telcobench_train} presents multiple-choice question answering results, where \textbf{GPT-4o} achieves the highest overall accuracy (85.6\%), surpassing other models in handling diverse telecom-related queries. Similarly, \textbf{Phi 4} and \textbf{Qwen2.5 72B} demonstrate strong performance, making them viable candidates for real-time network troubleshooting.

\begin{figure}[h]
    \centering
    \includegraphics[width=0.4\textwidth]{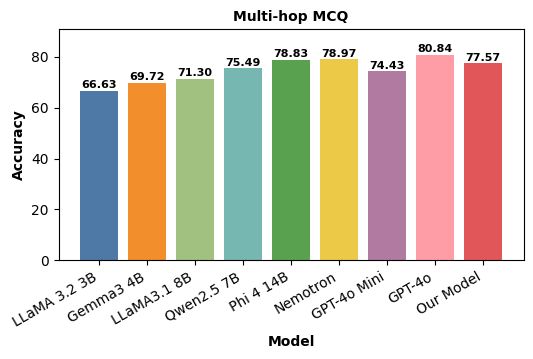}
    \caption{Multi-hop MCQ}
    \label{fig:multi_hop_mcq}
\end{figure}

For scenario-based filter generation from PCAP analysis (Table~\ref{tab:scenario_filter}), \textbf{phi-4-mini 3.8B} achieves the highest \textbf{LLM Judge Score} (66.63) and a strong \textbf{SEM Score} (0.8897), showcasing its effectiveness in generating relevant network filtering rules. \textbf{Llama 3.2 3B} follows closely with competitive ROUGE and SEM scores, indicating its ability to extract meaningful insights from network traffic data. \textbf{GPT-4o-mini} exhibits the highest \textbf{SEM Score} (0.8950) and excels in understanding complex PCAP structures, making it a robust choice for automated network security and traffic analysis.

\begin{figure}[h]
    \centering
    \includegraphics[width=0.3\textwidth]{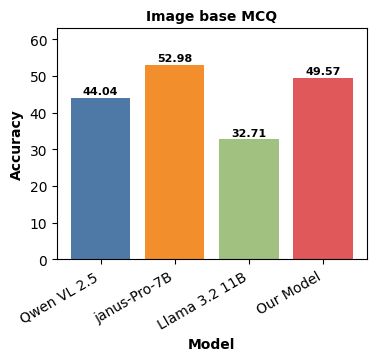}
    \caption{Image Based MCQ}
    \label{fig:image_based_mcq}
\end{figure}

\noindent\textbf{RQ3: Private Fine-Tuned LLMs for Telecom:}
To address RQ3, we evaluated the impact of fine-tuning telecom-specific LLMs on performance and data privacy. As shown in Table \ref{tab:mm_telcobench_train} and figure\ref{fig:multi_hop_mcq}, the fine-tuned Llama3.1 8B model significantly outperformed its original version on standard MCQs (84.58\% vs. 72.6\%) and multi-hop MCQs (77.57\% vs. 71.3\%), indicating enhanced understanding of telecom-specific knowledge. Similarly, figure \ref{fig:image_based_mcq} shows an improvement in the Qwen VL2.5 model's performance on Image-MCQs (49.57\% vs. 44.04\%), suggesting better integration of visual and textual reasoning. However, as reflected in Table \ref{tab:Llama3.1 finetune QA}, the improvements in Long-QA and Multi-hop QA tasks were relatively modest, indicating that complex reasoning in domain-specific contexts remains challenging. Overall, the results demonstrate that fine-tuning LLMs with telecom-specific data enhances their accuracy and relevance while maintaining data privacy, though further optimization may be needed for more complex tasks.

\noindent\textbf{RQ4: Text-to-Image Generation for Documentation:}
To address RQ4, we evaluated multiple AI models on their ability to generate precise and relevant images for technical documentation, troubleshooting guides, and other telecom-related visual content. The evaluation metrics included precision, recall, F1-score, accuracy, and coverage of domain-specific visual elements.
\textbf{GPT-4o-mini} demonstrated the highest performance across all metrics, achieving a perfect score of 1.0000 for precision, recall, F1-score, and accuracy, alongside a high domain coverage of 0.8295 reffered in table ~\ref{tab:packet_sequence} while our fine-tuned model \textbf{LLama-VL-Telco} matched the performance of GPT-4o-mini across most metrics . The fine-tuning process enhanced the model's precision and contextual accuracy, making it well-suited for telecom-specific scenarios. For the image  updation tasks,the model was able to update the image accurately as per the prompt given as shown in Figure~\ref{fig:image correct sequence}

\noindent\textbf{RQ5:Real-Time AI Agent Support:}
Our \textbf{RAG-based} multimodal ChatBot  was able to demonstrate efficient, real-time customer support by accurately managing inquiries and streamlining communication processes. By leveraging retrieval-augmented generation, the model retrieves relevant information,image from a large knowledge base, ensuring that responses are accurate and contextually appropriate. Additionally, the model’s ability to handle complex queries,provide quick resolutions and image interpretability(as shown in figure~\ref{fig:chatbot_multimodal}contributes to increased efficiency in customer service operations, resulting in a significant reduction in workload for support teams.

\begin{table*}[ht]
    \centering
    \fontsize{10pt}{8pt}\selectfont
    \renewcommand{\arraystretch}{1.1}
    \setlength{\tabcolsep}{1.5pt} 
    \begin{tabular}{lcccccccccccccccc}
        \toprule
        \textbf{Model} & 
        \makecell{\textbf{CT}\\\textbf{WG1}} & 
        \makecell{\textbf{CT}\\\textbf{WG3}} & 
        \makecell{\textbf{CT}\\\textbf{WG4}} & 
        \makecell{\textbf{CT}\\\textbf{WG6}} & 
        \makecell{\textbf{RAN}\\\textbf{1}} & 
        \makecell{\textbf{RAN}\\\textbf{2}} & 
        \makecell{\textbf{RAN}\\\textbf{3}} & 
        \makecell{\textbf{RAN}\\\textbf{4}} & 
        \makecell{\textbf{RAN}\\\textbf{5}} & 
        \makecell{\textbf{SA}\\\textbf{WG1}} & 
        \makecell{\textbf{SA}\\\textbf{WG2}} & 
        \makecell{\textbf{SA}\\\textbf{WG3}} & 
        \makecell{\textbf{SA}\\\textbf{WG4}} & 
        \makecell{\textbf{SA}\\\textbf{WG5}} & 
        \makecell{\textbf{SA}\\\textbf{WG6}} & 
        \makecell{\textbf{Acc-}\\\textbf{uracy}} \\
        \midrule
        Llama 3.2 3B       & 64.0  & 70.5 & 68.8 & 64.5 & 61.9 & 63.8 & 68.4 & 52.6 & 58.8   & 75.5 & 70.9 & 65.8 & 69.8 & 72.4 & 73.7 & 66.8 \\
        Llama 3.2 11B & 69.0  & 78.9 & 71.1 & 69.7 & 66.4 & 71.2 & 75.8 & 57.8& 64.8 & 78.3 & 77.1 & 73.9 & 75.8 & 75.3& 79.3 & 72.6  \\
        Llama 3.3 70B       & 80.6  & 85.7 & 83.6 & 80.5 & 75.4 & 82.0 & 83.5  & 70.3 & 73.0 & 86.8 & 84.6 & 81.5 & 83.4 & 86.2 & 85.4  & 81.9 \\
        Nemotron 70B            & 80.8  & 84.8 & 81.1  & 79.3 & 77.4 & 82.5  & 83.1 & 69.1 & 71.9 & 86.6 & 84.6 & 82.3 & 83.6 & 88.0 & 86.2 & 81.8 \\
        Qwen2.5 72B         & 79.9  & 86.4 & 81.8 & 79.5 & 78.0 & 84.0 & 85.3 & 68.9 & 74.8 & 85.7 & 85.1  & 82.8 & 84.2  & 88.2  & 85.6 & 82.4 \\
        GPT-4o              & 84.5  & 88.7  & 85.9  & 84.7 & 79.3 & 87.3 & 88.3 & 70.9 & 76.4 & 89.1 & 87.6 & 85.5 & 88.7 & 90.1 & 89.3 & \textbf{85.6} \\
        GPT-4o-mini         & 75.7  & 81.2  & 80.7 & 76.5 & 70.3 & 79.3 & 77.9 & 66.0 & 69.4 & 83.8 & 81.8 & 79.0 & 80.4 & 84.2 & 82.7 & 78.4 \\
        Phi 4               & 81.0   & 87.6  & 83.6  & 83.1  & 76.7  & 83.9  & 86.7  & 66.7  & 78.2  & 86.4  & 85.7  & 83.6  & 87.7  & 88.2  & 89.1  & 83.7 \\
        \textbf{Our model }              & 82.8   & 88.7  & 83.1  & 86.6  & 86.8  & 83.6  & 87.9  & 75.1  & 79.7  & 87.1  & 87.2  & 84.8  & 86.7  & 88.3  & 88.2  & 84.9 \\
        \bottomrule
    \end{tabular}
    \caption{Multiple-Choice Question Answer Evaluation Result}
    \label{tab:mm_telcobench_train}
\end{table*}

\begin{table}[ht]
    \centering
    \fontsize{10pt}{12pt}\selectfont
    \renewcommand{\arraystretch}{1.2}
    \setlength{\tabcolsep}{1pt} 
    \begin{tabular}{lccccc}
        \toprule
        \textbf{Model} & \makecell{\textbf{ROUGE} \\ \textbf{1}} & \makecell{\textbf{ROUGE} \\ \textbf{2}} & \makecell{\textbf{ROUGE} \\ \textbf{L}} & \begin{tabular}{@{}c@{}}\textbf{Sem} \\ \textbf{Score} \end{tabular} & \textbf{BLEU} \\
        \midrule
        LLama3.2 3B & 0.20 & 0.09 & 0.16 & 0.74 & 4.41 \\
        phi4 14B & 0.16 & 0.07 & 0.12 & 0.84 & 2.83 \\
        LLama3.1 8B & 0.24 & 0.11 & 0.19 & 0.84 & 5.72 \\
        Nemotron 70B & 0.15 & 0.07 & 0.11 & 0.84 & 2.56 \\
        Qwen2.5VL 7B & 0.27 & 0.11 & 0.20 & 0.83 & 5.65 \\
        GPT-4o & 0.19 & 0.09 & 0.15 & 0.85 & 4.36 \\
        \bottomrule
    \end{tabular}
    \caption{Long-Form Telecom Blog QA Results}
    \label{tab:long blog telecom}
\end{table}


\begin{table*}[h]
    \centering
    \fontsize{10pt}{12pt}\selectfont
    \renewcommand{\arraystretch}{1.1}
    \setlength{\tabcolsep}{4pt}
    \begin{tabular}{llll}
        \toprule
        \textbf{Model} & \textbf{Top1 Accuracy} & \textbf{Top3 Accuracy}& \textbf{Top5 Accuracy} \\
        \midrule
        all-MiniLM-L6-v2 & 0.665  & 0.817 & 0.858  \\
        all-mpnet-base-v2 & 0.635  & 0.784 & 0.822 \\
        bge-small-en-v1.5 & 0.706  & 0.827 & 0.861 \\
        bge-large-en-v1.5 & \textbf{0.769}  & \textbf{0.882} & \textbf{0.914}\\
        snowflake-arctic-embed-m & 0.476  & 0.59 & 0.626\\
        snowflake-arctic-embed-I & 0.392  & 0.498 & 0.537\\
        mxbai-embed-large-v1 & 0.755  & 0.87 & 0.895\\

        \bottomrule
    \end{tabular}
    \caption{Information Retrieval Evaluation Results}
    \label{tab:mm_telcobench_train}
\end{table*}

\begin{table*}[ht]
    \centering
    \fontsize{10pt}{12pt}\selectfont
    \renewcommand{\arraystretch}{1.2}
    \setlength{\tabcolsep}{2pt} 
    \begin{tabular}{lcccccc}
        \toprule
        \textbf{Model} & \makecell{\textbf{ROUGE} \\ \textbf{1}} & \makecell{\textbf{ROUGE} \\ \textbf{2}} & \makecell{\textbf{ROUGE} \\ \textbf{L}} & \textbf{SEM Score} & \makecell{\textbf{LLM Judge} \\ \textbf{Score}} & \textbf{sacreBLEU} \\
        \midrule
        Llama3.2 3B & 0.39 & 0.13 & 0.22 & 0.76 & 45.41 & 8.25 \\
        Phi 4 14B & 0.39 & 0.13 & 0.20 & 0.89 & 61.14 & 8.53 \\
        Llama3.1 8B & 0.39 & 0.12 & 0.20 & 0.89 & 60.20 & 8.29 \\
        Nemotron 70B & 0.38 & 0.12 & 0.20 & 0.90 & 72.04 & 6.58 \\
        Qwen2.5VL 7B & 0.28 & 0.11 & 0.18 & 0.88 & 43.23 & 3.40 \\
        GPT-4o & 0.41 & 0.15 & 0.23 & 0.91 & 75.39 & 10.00 \\
        \bottomrule
    \end{tabular}
    \caption{Long QA}
    \label{Long QA}
\end{table*}

\begin{table*}[h]
    \centering
    \fontsize{10pt}{12pt}\selectfont
    \renewcommand{\arraystretch}{1.2}
    \setlength{\tabcolsep}{2pt} 
    \begin{tabular}{lccccc}
        \toprule
        \textbf{Model Name} & \textbf{sacreBLEU} & \makecell{\textbf{ROUGE} \\ \textbf{L}} & \makecell{\textbf{ROUGE} \\ \textbf{1}} & \makecell{\textbf{ROUGE} \\ \textbf{2}} & \textbf{SEM Score} \\
        \midrule
        Qwen2.5-VL-7B-Instruct & 0.08 & 0.29 & 0.36 & 0.21 & 0.89 \\
        Llama-3.2-11B-Vision-Instruct & 0.09 & 0.06 & 0.34 & 0.21 & 0.89 \\
        Janus-Pro-7B & 0.13 & 0.32 & 0.38 & 0.25 & 0.84 \\
        \bottomrule
    \end{tabular}
    \caption{Image Based QA}
    \label{tab:model_comparison_final}
\end{table*}

\begin{table*}[h]
    \centering
    \fontsize{10pt}{12pt}\selectfont
    \renewcommand{\arraystretch}{1.2}
    \setlength{\tabcolsep}{2pt} 
    \begin{tabular}{lccccc}
        \toprule
        \textbf{Model Name} & \textbf{sacreBLEU} & \makecell{\textbf{ROUGE} \\ \textbf{L}} & \makecell{\textbf{ROUGE} \\ \textbf{1}} & \makecell{\textbf{ROUGE} \\ \textbf{2}} & \textbf{SEM Score} \\
        \midrule
        Qwen2.5-VL-7B-Instruct & 2.06 & 0.15 & 0.18 & 0.04 & 0.73 \\
        Llama-3.2-11B-Vision-Instruct & 9.51 & 0.28 & 0.07 & 0.01 & 0.67 \\
        Janus-Pro-7B & 0.56 & 0.06 & 0.07 & 0.01 & 0.68 \\
        \bottomrule
    \end{tabular}
    \caption{Image caption generation.}
    \label{tab:image caption}
\end{table*}

\begin{table*}[h]
    \centering
    \fontsize{10pt}{12pt}\selectfont
    \renewcommand{\arraystretch}{1.2}
    \setlength{\tabcolsep}{2pt} 
    \begin{tabular}{lcccc}
        \toprule
        \textbf{Model} & \makecell{\textbf{Top 1} \\ \textbf{Accuracy}} & \makecell{\textbf{Top 3} \\ \textbf{Accuracy}} & \makecell{\textbf{Top 5} \\ \textbf{Accuracy}} & \makecell{\textbf{Top 10} \\ \textbf{Accuracy}} \\
        \midrule
        ColPali & 0.42 & 0.66 & 0.72 & 0.84 \\
        Clip Based (laion/CLIP-ViT-H-14-laion2B-s32B-b79K) & 0.24 & 0.36 & 0.44 & 0.50 \\
        \bottomrule
    \end{tabular}
    \caption{Image retrieval.}
    \label{tab:image retrieval}
\end{table*}

\begin{table*}[h]
    \centering
    \fontsize{10pt}{12pt}\selectfont
    \renewcommand{\arraystretch}{1.2}
    \setlength{\tabcolsep}{2pt} 
    \begin{tabular}{lc}
        \toprule
        \textbf{Model} & \textbf{Accuracy} \\
        \midrule
        Qwen VL 2.5 & 44.04 \\
        Llama 3.2 11B & 32.71 \\
        Janus-Pro-7B & 52.98 \\
        \bottomrule
    \end{tabular}
    \caption{Image Based MCQ.}
    \label{tab:image MCQ}
\end{table*}

\begin{table*}[h]
    \centering
    \fontsize{10pt}{12pt}\selectfont
    \renewcommand{\arraystretch}{1.2}
    \setlength{\tabcolsep}{2pt} 
    \begin{tabular}{lccccc}
        \toprule
        \textbf{Task} & \textbf{sacreBLEU} & \makecell{\textbf{ROUGE} \\ \textbf{L}} & \makecell{\textbf{ROUGE} \\ \textbf{1}} & \makecell{\textbf{ROUGE} \\ \textbf{2}} & \textbf{SEM Score} \\
        \midrule
        Long-QA & 9.33& 0.19 & 0.40 & 0.13 & 0.88 \\
        Multi-hop QA & 6.62 & 0.12 & 0.20 & 0.07 & 0.80 \\
        \bottomrule
    \end{tabular}
    \caption{Llama3.1 8B result on Long type QA}
    \label{tab:Llama3.1 finetune QA}
\end{table*}

\begin{table*}[h]
    \centering
    \small
    \renewcommand{\arraystretch}{1.2}
    \setlength{\tabcolsep}{1.2pt}
    \begin{tabular}{llccccccc}
        \toprule
        \textbf{Diagram Type} & \textbf{Model} &\textbf{BLEU} & \textbf{SacreBleu} & \textbf{CodeBleu} & \textbf{Meteor} & \textbf{chrF} & \textbf{Bleurt} & \textbf{Rouge-L} \\
        \midrule
        \multirow{7}{*}{\textbf{Sequence}}
        & Llama3.2-11B-Instruct(vanilla)    & 0.7491 & 74.9058 & 0.7491 & 0.8453 & 94.9214 & 0.8799 & 0.9911\\
        & Qwen8B  & 0.7320 & 73.2011 & 0.7320 & 0.8325 & 94.6114 & 0.8642 & 0.9875  \\
        & miniCPM     & 0.5688 & 56.9839 & 0.5688 & 0.7226 & 84.4162 & 0.6575 & 0.8780  \\
        & GPT-4o-mini     & 0.8511 & 85.1115 & 0.8511 & 0.8815 & 96.7221 & 0.8917 & 0.9997 \\
        & \textbf{Llama3.2-11B-Instruct-finetuned}& \textbf{1.0000} & \textbf{100.0000} & \textbf{1.0000} & \textbf{0.9999} & \textbf{100.0000} & \textbf{1.0068} & \textbf{1.0000}\\
        
        \midrule
        \multirow{7}{*}{\textbf{Packet}}
        & Llama3.2-11B-Instruct(vanilla) & 0.9996 & 99.9555 & 0.9996 & 1.0000 & 99.9915 & 0.8249 & 0.9999\\
        & Qwen8B   & 0.9696 & 96.9629 & 0.9696 & 0.9992 & 97.5036 & 0.7980 & 0.9791  \\
        & miniCPM     &  0.9122 & 91.2241 & 0.9122 & 0.9513 & 95.2885 & 0.7540 & 0.9548  \\
        & GPT-4o-mini     &  \textbf{1.0000} & \textbf{100.0000} & \textbf{1.0000} & \textbf{1.0000} & \textbf{100.0000} & \textbf{0.8312} & \textbf{1.0000} \\
        & \textbf{Llama3.2-11B-Instruct-finetuned }& \textbf{1.0000} & \textbf{100.0000} & \textbf{1.0000} & \textbf{1.0000} & \textbf{100.0000} & \textbf{0.8355} & \textbf{1.0000}\\
        
        \bottomrule
    \end{tabular}
    \caption{ results for Packet and Sequence diagram types.}
    \label{tab:packet_sequence}
\end{table*}

\begin{table*}[h]
    \centering
    \small
    \renewcommand{\arraystretch}{1.2}
    \setlength{\tabcolsep}{1.2pt}
\begin{tabular}{lcccccc}
\hline
\textbf{Model} & \textbf{ROUGE 1} & \textbf{ROUGE 2} & \textbf{ROUGE L} & \textbf{SEM Score} & \textbf{LLM Judge Score} & \textbf{SacreBLEU} \\
\hline
\textbf{phi-4-mini 3.8B} & 0.1867 & 0.0787 & 0.1354 & 0.8897 & 66.63 & 4.07 \\
\textbf{llama 3.2 3B} & 0.1898 & 0.0760 & 0.1398 & 0.8838 & 61.03 & 3.49 \\
\textbf{gpt-4o-mini} & 0.1787 & 0.0746 & 0.1273 & 0.8950 & 73.53 & 3.28 \\
\textbf{Gemma3 4B} & 0.1019 & 0.0367 & 0.0715 & 0.8763 & 61.53 & 1.33 \\
\hline
\end{tabular}
\caption{Performance Comparison of Various Models on Scenario-based Filter Generation Benchmark}
\label{tab:scenario_filter}
\end{table*}
\section{Discussions, Perspectives, and Future Directions}

In this paper, we proposed \benchmarkname\ to promote multi-modal LLM research and development for the telecom domain. We cover a wide range of tasks that could be of value to Telecom industries, but don't attempt to make it exhaustive. Nevertheless, we plan to keep updating \benchmarkname\ as we get more inputs from the community. Multimodal LLMs are an active area of research with new versions frequently surpassing the previous ones. A model ranked today as the most suitable for a specific task may become obsolete within weeks or months, calling for continuous and periodically updated benchmarks and evaluations. We believe that the techniques presented in this paper would be very useful for the same. Further, advanced techniques based on reinforcement learning could be used to build upon the work in this paper. 

The ongoing work within 3GPP community is focused on refining and enhancing communication standards that support emerging technologies such as 5G Advanced and the transition to 6G. Our research aligns perfectly with current needs, add depth and new dimensions to this evolving landscape, and foster AI-assisted standardization workflows in 6G and beyond. This ensures timely standardization activities to maintain the strict timelines for various releases, very crucial to respect market launch commitments and expected return on investment (ROI). Our structured domain-specific evaluation and benchmark framework can efficiently assist in this incremental and iterative development process, through features such as the support of the multihop MCQs, ensuring cross-release coherence. 

Further, our framework can play a pivotal role in fostering close collaboration between telecom stakeholders, aligning industry priorities and preventing conflicting specifications. Since 3GPP is continuously coordinating its technical work with external standard development organizations (SDOs) to ensure harmonization between various standardization efforts (e.g. using Liaison Statements (LSs) exchanges), our framework could be replicated for other telecom standardization bodies such as International Telecommunication Union (ITU) and Institute of Electrical and Electronics Engineers (IEEE), leading to a broader automated and AI-driven transformation in global telecom standardization ecosystem. Our framework can also evolve to go beyond text and image to include audio as well as video data repositories anticipated to be very useful for analyzing live meeting discussions and debates from various 3GPP working groups, processing spoken contributions in real-time, automating the generation of comprehensive meeting summaries, and detecting inconsistencies and redundancies across the presented proposals (e.g. technical documents (Tdocs)). 

3GPP standards cover a wide range of technical domains, including protocols, architectures, and interfaces. A multimodal benchmark suite may struggle to provide equally efficient insights for every 3GPP WG, as can be observed in Table~\ref{tab:mm_telcobench_train}. Moreover, 3GPP standards contain specifications, technical reports, meeting notes, contributions, and voting mechanisms, each with different formats and complexity \cite{maatouk2024largelanguagemodelstelecom} which require additional fine-tuning. Furthermore, LLMs may lack explainability, making it difficult for 3GPP players to verify and trust their recommendations with the risk of adding more burden and manual checks instead of the promise of reducing it, requiring explainable AI tools to be deployed in parallel for understandable and transparent findings.

\section{Conclusion} \label{sec:conclusion}

In this paper, we introduce \benchmarkname, \textbf{a comprehensive suite of multimodal benchmarks and models} tailored for the telecom domain. The benchmark introduces various tasks (both text based and image based) that are motivated by industrial automation use cases such as network operations, network management, improving documentation quality, and retrieval of relevant text and images. Our experiments show that the open source LLMs such as Llama 3.2 11B can achieve competitive performance as compared to ChatGPT after fine-tuning for several tasks. We observe that Image-based tasks are more challenging for the current generation of VLMs. We conclude that the benchmarks and use cases discussed in this paper can help the telecom sector achieve enhanced operational efficiency, cost savings, and superior customer service.

\bibliographystyle{ACM-Reference-Format}
\bibliography{references}

@inproceedings{yuan2024continued,
  title={A continued pretrained llm approach for automatic medical note generation},
  author={Yuan, Dong and Rastogi, Eti and Naik, Gautam and Rajagopal, Sree Prasanna and Goyal, Sagar and Zhao, Fen and Chintagunta, Bharath and Ward, Jeffrey},
  booktitle={Proceedings of the 2024 Conference of the North American Chapter of the Association for Computational Linguistics: Human Language Technologies (Volume 2: Short Papers)},
  pages={565--571},
  year={2024}
}

@article{singhal2023large,
  title={Large language models encode clinical knowledge},
  author={Singhal, Karan and Azizi, Shekoofeh and Tu, Tao and Mahdavi, S Sara and Wei, Jason and Chung, Hyung Won and Scales, Nathan and Tanwani, Ajay and Cole-Lewis, Heather and Pfohl, Stephen and others},
  journal={Nature},
  volume={620},
  number={7972},
  pages={172--180},
  year={2023},
  publisher={Nature Publishing Group}
}

@article{didolkar2024metacognitive,
  title={Metacognitive capabilities of llms: An exploration in mathematical problem solving},
  author={Didolkar, Aniket and Goyal, Anirudh and Ke, Nan Rosemary and Guo, Siyuan and Valko, Michal and Lillicrap, Timothy and Jimenez Rezende, Danilo and Bengio, Yoshua and Mozer, Michael C and Arora, Sanjeev},
  journal={Advances in Neural Information Processing Systems},
  volume={37},
  pages={19783--19812},
  year={2024}
}

@inproceedings{satpute2024can,
  title={Can llms master math? investigating large language models on math stack exchange},
  author={Satpute, Ankit and Gie{\ss}ing, Noah and Greiner-Petter, Andr{\'e} and Schubotz, Moritz and Teschke, Olaf and Aizawa, Akiko and Gipp, Bela},
  booktitle={Proceedings of the 47th international ACM SIGIR conference on research and development in information retrieval},
  pages={2316--2320},
  year={2024}
}

@article{roziere2023code,
  title={Code llama: Open foundation models for code},
  author={Roziere, Baptiste and Gehring, Jonas and Gloeckle, Fabian and Sootla, Sten and Gat, Itai and Tan, Xiaoqing Ellen and Adi, Yossi and Liu, Jingyu and Sauvestre, Romain and Remez, Tal and others},
  journal={arXiv preprint arXiv:2308.12950},
  year={2023}
}

@article{lin2023pushing,
  title={Pushing large language models to the 6g edge: Vision, challenges, and opportunities},
  author={Lin, Zheng and Qu, Guanqiao and Chen, Qiyuan and Chen, Xianhao and Chen, Zhe and Huang, Kaibin},
  journal={arXiv preprint arXiv:2309.16739},
  year={2023}
}

@inproceedings{lee2024telbench,
  title={TelBench: A Benchmark for Evaluating Telco-Specific Large Language Models},
  author={Lee, Sunwoo and Arya, Dhammiko and Cho, Seung-Mo and Han, Gyoung-eun and Hong, Seokyoung and Jang, Wonbeom and Lee, Seojin and Park, Sohee and Sek, Sereimony and Song, Injee and others},
  booktitle={Proceedings of the 2024 Conference on Empirical Methods in Natural Language Processing: Industry Track},
  pages={609--626},
  year={2024}
}

@inproceedings{10.1145/3636534.3649379,
author = {Wen, Hao and Li, Yuanchun and Liu, Guohong and Zhao, Shanhui and Yu, Tao and Li, Toby Jia-Jun and Jiang, Shiqi and Liu, Yunhao and Zhang, Yaqin and Liu, Yunxin},
title = {AutoDroid: LLM-powered Task Automation in Android},
year = {2024},
isbn = {9798400704895},
publisher = {Association for Computing Machinery},
address = {New York, NY, USA},
url = {https://doi.org/10.1145/3636534.3649379},
doi = {10.1145/3636534.3649379},
booktitle = {Proceedings of the 30th Annual International Conference on Mobile Computing and Networking},
pages = {543–557},
numpages = {15},
location = {Washington D.C., DC, USA},
series = {ACM MobiCom '24}
}

@article{luo2023wizardmath,
  title={Wizardmath: Empowering mathematical reasoning for large language models via reinforced evol-instruct},
  author={Luo, Haipeng and Sun, Qingfeng and Xu, Can and Zhao, Pu and Lou, Jianguang and Tao, Chongyang and Geng, Xiubo and Lin, Qingwei and Chen, Shifeng and Zhang, Dongmei},
  journal={arXiv preprint arXiv:2308.09583},
  year={2023}
}

@inproceedings{luukkonen-etal-2023-fingpt,
    title = "{F}in{GPT}: Large Generative Models for a Small Language",
    author = "Luukkonen, Risto  and
      Komulainen, Ville  and
      Luoma, Jouni  and
      Eskelinen, Anni  and
      Kanerva, Jenna  and
      Kupari, Hanna-Mari  and
      Ginter, Filip  and
      Laippala, Veronika  and
      Muennighoff, Niklas  and
      Piktus, Aleksandra  and
      Wang, Thomas  and
      Tazi, Nouamane  and
      Scao, Teven  and
      Wolf, Thomas  and
      Suominen, Osma  and
      Sairanen, Samuli  and
      Merioksa, Mikko  and
      Heinonen, Jyrki  and
      Vahtola, Aija  and
      Antao, Samuel  and
      Pyysalo, Sampo",
    booktitle = "Proceedings of the 2023 Conference on Empirical Methods in Natural Language Processing",
    month = dec,
    year = "2023",
    address = "Singapore",
    publisher = "Association for Computational Linguistics",
    url = "https://aclanthology.org/2023.emnlp-main.164/",
    doi = "10.18653/v1/2023.emnlp-main.164",
    pages = "2710--2726"
}

@inproceedings{yao2024lawyer,
  title={Lawyer GPT: A legal large language model with enhanced domain knowledge and reasoning capabilities},
  author={Yao, Shunyu and Ke, Qingqing and Wang, Qiwei and Li, Kangtong and Hu, Jie},
  booktitle={Proceedings of the 2024 3rd International Symposium on Robotics, Artificial Intelligence and Information Engineering},
  pages={108--112},
  year={2024}
}

@misc{zou2024telecomgptframeworkbuildtelecomspecfic,
      title={TelecomGPT: A Framework to Build Telecom-Specfic Large Language Models}, 
      author={Hang Zou and Qiyang Zhao and Yu Tian and Lina Bariah and Faouzi Bader and Thierry Lestable and Merouane Debbah},
      year={2024},
      eprint={2407.09424},
      archivePrefix={arXiv},
      primaryClass={eess.SP},
      url={https://arxiv.org/abs/2407.09424}, 
}

@misc{nikbakht2024tspecllmopensourcedatasetllm,
      title={TSpec-LLM: An Open-source Dataset for LLM Understanding of 3GPP Specifications}, 
      author={Rasoul Nikbakht and Mohamed Benzaghta and Giovanni Geraci},
      year={2024},
      eprint={2406.01768},
      archivePrefix={arXiv},
      primaryClass={cs.NI},
      url={https://arxiv.org/abs/2406.01768}, 
}

@misc{bariah2023understandingtelecomlanguagelarge,
      title={Understanding Telecom Language Through Large Language Models}, 
      author={Lina Bariah and Hang Zou and Qiyang Zhao and Belkacem Mouhouche and Faouzi Bader and Merouane Debbah},
      year={2023},
      eprint={2306.07933},
      archivePrefix={arXiv},
      primaryClass={cs.CL},
      url={https://arxiv.org/abs/2306.07933}, 
}

@inproceedings {298168,
author = {Mirza Masfiqur Rahman and Imtiaz Karim and Elisa Bertino},
title = {{CellularLint}: A Systematic Approach to Identify Inconsistent Behavior in Cellular Network Specifications},
booktitle = {33rd USENIX Security Symposium (USENIX Security 24)},
year = {2024},
isbn = {978-1-939133-44-1},
address = {Philadelphia, PA},
pages = {5215--5232},
url = {https://www.usenix.org/conference/usenixsecurity24/presentation/rahman},
publisher = {USENIX Association},
month = aug
}

@inproceedings{karim-etal-2023-spec5g,
    title = "{SPEC}5{G}: A Dataset for 5{G} Cellular Network Protocol Analysis",
    author = "Karim, Imtiaz  and
      Mubasshir, Kazi Samin  and
      Rahman, Mirza Masfiqur  and
      Bertino, Elisa",
    editor = "Park, Jong C.  and
      Arase, Yuki  and
      Hu, Baotian  and
      Lu, Wei  and
      Wijaya, Derry  and
      Purwarianti, Ayu  and
      Krisnadhi, Adila Alfa",
    booktitle = "Findings of the Association for Computational Linguistics: IJCNLP-AACL 2023 (Findings)",
    month = nov,
    year = "2023",
    address = "Nusa Dua, Bali",
    publisher = "Association for Computational Linguistics",
    url = "https://aclanthology.org/2023.findings-ijcnlp.3/",
    doi = "10.18653/v1/2023.findings-ijcnlp.3",
    pages = "20--38"
}

@misc{wang2023makingnetworkconfigurationhuman,
      title={Making Network Configuration Human Friendly}, 
      author={Changjie Wang and Mariano Scazzariello and Alireza Farshin and Dejan Kostic and Marco Chiesa},
      year={2023},
      eprint={2309.06342},
      archivePrefix={arXiv},
      primaryClass={cs.NI},
      url={https://arxiv.org/abs/2309.06342}, 
}

@inproceedings{10.1145/3636534.3649380,
author = {Sun, Chuanhao and Pawar, Ujjwal and Khoja, Molham and Foukas, Xenofon and Marina, Mahesh K. and Radunovic, Bozidar},
title = {SpotLight: Accurate, Explainable and Efficient Anomaly Detection for Open RAN},
year = {2024},
isbn = {9798400704895},
publisher = {Association for Computing Machinery},
address = {New York, NY, USA},
url = {https://doi.org/10.1145/3636534.3649380},
doi = {10.1145/3636534.3649380},
booktitle = {Proceedings of the 30th Annual International Conference on Mobile Computing and Networking},
pages = {923–937},
numpages = {15},
location = {Washington D.C., DC, USA},
series = {ACM MobiCom '24}
}

@article{Yang2023HarnessingTP,
  title={Harnessing the Power of LLMs in Practice: A Survey on ChatGPT and Beyond},
  author={Jingfeng Yang and Hongye Jin and Ruixiang Tang and Xiaotian Han and Qizhang Feng and Haoming Jiang and Bing Yin and Xia Hu},
  journal={ACM Transactions on Knowledge Discovery from Data},
  year={2023},
  volume={18},
  pages={1 - 32},
  url={https://api.semanticscholar.org/CorpusID:258331833}
}

@misc{commoncrawl,
  author = "{Common Crawl}",
  title = "{Common Crawl Dataset}",
  year = {2024},
  url = {https://commoncrawl.org/}
}

@misc{openwebtext,
  author = "{OpenWebText}",
  title = "{OpenWebText Corpus}",
  year = {2024},
  url = {https://skylion007.github.io/OpenWebTextCorpus/}
}

@inproceedings{10.1145/3703323.3704278,
author = {Kumar, Anshul and Gupta, Gagan Raj and Kumar, Sunny and Sahu, Ankita and Chakraborty, Apu},
title = {What does your Packet Capture say?},
year = {2025},
isbn = {9798400711244},
publisher = {Association for Computing Machinery},
address = {New York, NY, USA},
url = {https://doi.org/10.1145/3703323.3704278},
doi = {10.1145/3703323.3704278},
booktitle = {Proceedings of the 8th International Conference on Data Science and Management of Data (12th ACM IKDD CODS and 30th COMAD)},
pages = {427–431},
numpages = {5},
location = {
},
series = {CODS-COMAD '24}
}

@article{Kuckreja2023GeoChat:Grounded,title={GeoChat:Grounded Large Vision-Language Model for Remote Sensing},author={Kartik Kuckreja and M. S. Danish and Muzammal Naseer and Abhijit Das and Salman Khan and F. Khan},journal={2024 IEEE/CVF Conference on Computer Vision and Pattern Recognition (CVPR)},year={2023},pages={27831-27840},doi={10.1109/CVPR52733.2024.02629}}

@article{Zhou2023VisionLM,
  title={Vision Language Models in Autonomous Driving: A Survey and Outlook},
  author={Xingcheng Zhou and Mingyu Liu and Ekim Yurtsever and Bare Luka Žagar and Walter Zimmer and Hu Cao and Alois C. Knoll},
  journal={IEEE Transactions on Intelligent Vehicles},
  year={2023},
  url={https://api.semanticscholar.org/CorpusID:269865211}
}

@article{Hartsock2024Vision-Language,
title={Vision-Language Models for Medical Report Generation and Visual Question Answering: A Review},
author={Iryna Hartsock and Ghulam Rasool},
journal={ArXiv},
year={2024},
volume={abs/2403.02469},
doi={10.48550/arXiv.2403.02469}
}

@misc{karapantelakis2024usinglargelanguagemodels,
      title={Using Large Language Models to Understand Telecom Standards}, 
      author={Athanasios Karapantelakis and Mukesh Thakur and Alexandros Nikou and Farnaz Moradi and Christian Orlog and Fitsum Gaim and Henrik Holm and Doumitrou Daniil Nimara and Vincent Huang},
      year={2024},
      eprint={2404.02929},
      archivePrefix={arXiv},
      primaryClass={cs.CL},
      url={https://arxiv.org/abs/2404.02929}, 
}

@misc{3gpp_releases,
  author = "{3rd Generation Partnership Project (3GPP)}",
  title = "{3GPP Specifications and Technologies - Releases}",
  year = {2024},
  url = {https://www.3gpp.org/specifications-technologies/releases}
}

@misc{wu2023bloomberggptlargelanguagemodel,
      title={BloombergGPT: A Large Language Model for Finance}, 
      author={Shijie Wu and Ozan Irsoy and Steven Lu and Vadim Dabravolski and Mark Dredze and Sebastian Gehrmann and Prabhanjan Kambadur and David Rosenberg and Gideon Mann},
      year={2023},
      eprint={2303.17564},
      archivePrefix={arXiv},
      primaryClass={cs.LG},
      url={https://arxiv.org/abs/2303.17564}, 
}

@misc{shao2024wirelessllmempoweringlargelanguage,
      title={WirelessLLM: Empowering Large Language Models Towards Wireless Intelligence}, 
      author={Jiawei Shao and Jingwen Tong and Qiong Wu and Wei Guo and Zijian Li and Zehong Lin and Jun Zhang},
      year={2024},
      eprint={2405.17053},
      archivePrefix={arXiv},
      primaryClass={cs.NI},
      url={https://arxiv.org/abs/2405.17053}, 
}

@misc{lam2024largelanguagemodelsplant,
      title={Large Language Models in Plant Biology}, 
      author={Hilbert Yuen In Lam and Xing Er Ong and Marek Mutwil},
      year={2024},
      eprint={2401.02789},
      archivePrefix={arXiv},
      primaryClass={q-bio.GN},
      url={https://arxiv.org/abs/2401.02789}, 
}

@misc{roychowdhury2024understandingdomainadaptedsentence,
      title={Towards Understanding Domain Adapted Sentence Embeddings for Document Retrieval}, 
      author={Sujoy Roychowdhury and Sumit Soman and H. G. Ranjani and Vansh Chhabra and Neeraj Gunda and Shashank Gautam and Subhadip Bandyopadhyay and Sai Krishna Bala},
      year={2024},
      eprint={2406.12336},
      archivePrefix={arXiv},
      primaryClass={cs.CL},
      url={https://arxiv.org/abs/2406.12336}, 
}

@misc{maatouk2023teleqnabenchmarkdatasetassess,
      title={TeleQnA: A Benchmark Dataset to Assess Large Language Models Telecommunications Knowledge}, 
      author={Ali Maatouk and Fadhel Ayed and Nicola Piovesan and Antonio De Domenico and Merouane Debbah and Zhi-Quan Luo},
      year={2023},
      eprint={2310.15051},
      archivePrefix={arXiv},
      primaryClass={cs.IT},
      url={https://arxiv.org/abs/2310.15051}, 
}

@misc{bai2023qwenvlversatilevisionlanguagemodel,
      title={Qwen-VL: A Versatile Vision-Language Model for Understanding, Localization, Text Reading, and Beyond}, 
      author={Jinze Bai and Shuai Bai and Shusheng Yang and Shijie Wang and Sinan Tan and Peng Wang and Junyang Lin and Chang Zhou and Jingren Zhou},
      year={2023},
      eprint={2308.12966},
      archivePrefix={arXiv},
      primaryClass={cs.CV},
      url={https://arxiv.org/abs/2308.12966}, 
}

@misc{bordes2024introductionvisionlanguagemodeling,
      title={An Introduction to Vision-Language Modeling}, 
      author={Florian Bordes and Richard Yuanzhe Pang and Anurag Ajay and Alexander C. Li and Adrien Bardes and Suzanne Petryk and Oscar Mañas and Zhiqiu Lin and Anas Mahmoud and Bargav Jayaraman and Mark Ibrahim and Melissa Hall and Yunyang Xiong and Jonathan Lebensold and Candace Ross and Srihari Jayakumar and Chuan Guo and Diane Bouchacourt and Haider Al-Tahan and Karthik Padthe and Vasu Sharma and Hu Xu and Xiaoqing Ellen Tan and Megan Richards and Samuel Lavoie and Pietro Astolfi and Reyhane Askari Hemmat and Jun Chen and Kushal Tirumala and Rim Assouel and Mazda Moayeri and Arjang Talattof and Kamalika Chaudhuri and Zechun Liu and Xilun Chen and Quentin Garrido and Karen Ullrich and Aishwarya Agrawal and Kate Saenko and Asli Celikyilmaz and Vikas Chandra},
      year={2024},
      eprint={2405.17247},
      archivePrefix={arXiv},
      primaryClass={cs.LG},
      url={https://arxiv.org/abs/2405.17247}, 
}

@misc{bornea2024telcoragnavigatingchallengesretrievalaugmented,
      title={Telco-RAG: Navigating the Challenges of Retrieval-Augmented Language Models for Telecommunications}, 
      author={Andrei-Laurentiu Bornea and Fadhel Ayed and Antonio De Domenico and Nicola Piovesan and Ali Maatouk},
      year={2024},
      eprint={2404.15939},
      archivePrefix={arXiv},
      primaryClass={cs.IR},
      url={https://arxiv.org/abs/2404.15939}, 
}

@ARTICLE{10794684,
  author={Said, Azzedine Idir Ait and Mekrache, Abdelkader and Boutiba, Karim and Ramantas, Kostas and Ksentini, Adlen and Rahmani, Moufida},
  journal={IEEE Transactions on Cognitive Communications and Networking}, 
  title={5G INSTRUCT Forge: An Advanced Data Engineering Pipeline for Making LLMs Learn 5G}, 
  year={2024},
  volume={},
  number={},
  pages={1-1},
  doi={10.1109/TCCN.2024.3516055}}

@misc{maatouk2024telellmsseriesspecializedlarge,
      title={Tele-LLMs: A Series of Specialized Large Language Models for Telecommunications}, 
      author={Ali Maatouk and Kenny Chirino Ampudia and Rex Ying and Leandros Tassiulas},
      year={2024},
      eprint={2409.05314},
      archivePrefix={arXiv},
      primaryClass={cs.IT},
      url={https://arxiv.org/abs/2409.05314}, 
}

@inproceedings{Soman_2023, series={AIMLSystems 2023},
   title={Observations on LLMs for Telecom Domain: Capabilities and Limitations},
   url={http://dx.doi.org/10.1145/3639856.3639892},
   DOI={10.1145/3639856.3639892},
   booktitle={The Third International Conference on Artificial Intelligence and Machine Learning Systems},
   publisher={ACM},
   author={Soman, Sumit and H. G., Ranjani},
   year={2023},
   month=oct, pages={1–5},
   collection={AIMLSystems 2023} }

@INPROCEEDINGS{10624781,
  author={Bian, Jie and Welzl, Michael and Kutuzov, Andrey and Arefyev, Nikolay},
  booktitle={2024 IEEE International Conference on Machine Learning for Communication and Networking (ICMLCN)}, 
  title={Tell Me Why: Language Models Help Explain the Rationale Behind Internet Protocol Design}, 
  year={2024},
  volume={},
  number={},
  pages={447-453},
  doi={10.1109/ICMLCN59089.2024.10624781}}

@misc{maatouk2024largelanguagemodelstelecom,
      title={Large Language Models for Telecom: Forthcoming Impact on the Industry}, 
      author={Ali Maatouk and Nicola Piovesan and Fadhel Ayed and Antonio De Domenico and Merouane Debbah},
      year={2024},
      eprint={2308.06013},
      archivePrefix={arXiv},
      primaryClass={cs.IT},
      url={https://arxiv.org/abs/2308.06013}, 
}

@article{YAO2024100211,
title = {A survey on large language model (LLM) security and privacy: The Good, The Bad, and The Ugly},
journal = {High-Confidence Computing},
volume = {4},
number = {2},
pages = {100211},
year = {2024},
issn = {2667-2952},
doi = {https://doi.org/10.1016/j.hcc.2024.100211},
url = {https://www.sciencedirect.com/science/article/pii/S266729522400014X},
author = {Yifan Yao and Jinhao Duan and Kaidi Xu and Yuanfang Cai and Zhibo Sun and Yue Zhang}
}

@article{Lior2024SEAM:,
title={SEAM: A Stochastic Benchmark for Multi-Document Tasks},
author={Gili Lior and Avi Caciularu and Arie Cattan and Shahar Levy and Ori Shapira and Gabriel Stanovsky},
journal={ArXiv},
year={2024},
volume={abs/2406.16086},
doi={10.48550/arXiv.2406.16086}
}

@misc{yu2024unleashingmultihopreasoningpotential,
      title={Unleashing Multi-Hop Reasoning Potential in Large Language Models through Repetition of Misordered Context}, 
      author={Sangwon Yu and Ik-hwan Kim and Jongyoon Song and Saehyung Lee and Junsung Park and Sungroh Yoon},
      year={2024},
      eprint={2410.07103},
      archivePrefix={arXiv},
      primaryClass={cs.CL},
      url={https://arxiv.org/abs/2410.07103}, 
}

@article{Zhang2023Vision-Language,
title={Vision-Language Models for Vision Tasks: A Survey},
author={Jingyi Zhang and Jiaxing Huang and Sheng Jin and Shijian Lu},
journal={IEEE Transactions on Pattern Analysis and Machine Intelligence},
year={2023},
volume={46},
pages={5625-5644},
doi={10.1109/TPAMI.2024.3369699}
}

@inproceedings{fei-etal-2025-internlm,
    title = "{I}ntern{LM}-Law: An Open-Sourced {C}hinese Legal Large Language Model",
    author = "Fei, Zhiwei  and
      Zhang, Songyang  and
      Shen, Xiaoyu  and
      Zhu, Dawei  and
      Wang, Xiao  and
      Ge, Jidong  and
      Ng, Vincent",
    editor = "Rambow, Owen  and
      Wanner, Leo  and
      Apidianaki, Marianna  and
      Al-Khalifa, Hend  and
      Eugenio, Barbara Di  and
      Schockaert, Steven",
    booktitle = "Proceedings of the 31st International Conference on Computational Linguistics",
    month = jan,
    year = "2025",
    address = "Abu Dhabi, UAE",
    publisher = "Association for Computational Linguistics",
    url = "https://aclanthology.org/2025.coling-main.629/",
    pages = "9376--9392"
}

@article{Wu2024PMC-LLaMA,
title={PMC-LLaMA: toward building open-source language models for medicine},
author={Chaoyi Wu and Weixiong Lin and Xiaoman Zhang and Ya Zhang and Weidi Xie and Yanfeng Wang},
journal={Journal of the American Medical Informatics Association : JAMIA},
year={2024},
doi={10.1093/jamia/ocae045}
}

@misc{hu2021loralowrankadaptationlarge,
      title={LoRA: Low-Rank Adaptation of Large Language Models}, 
      author={Edward J. Hu and Yelong Shen and Phillip Wallis and Zeyuan Allen-Zhu and Yuanzhi Li and Shean Wang and Lu Wang and Weizhu Chen},
      year={2021},
      eprint={2106.09685},
      archivePrefix={arXiv},
      primaryClass={cs.CL},
      url={https://arxiv.org/abs/2106.09685}, 
}

@misc{xu2023criticalevaluationevaluationslongform,
      title={A Critical Evaluation of Evaluations for Long-form Question Answering}, 
      author={Fangyuan Xu and Yixiao Song and Mohit Iyyer and Eunsol Choi},
      year={2023},
      eprint={2305.18201},
      archivePrefix={arXiv},
      primaryClass={cs.CL},
      url={https://arxiv.org/abs/2305.18201}, 
}

@misc{radford2021learningtransferablevisualmodels,
      title={Learning Transferable Visual Models From Natural Language Supervision}, 
      author={Alec Radford and Jong Wook Kim and Chris Hallacy and Aditya Ramesh and Gabriel Goh and Sandhini Agarwal and Girish Sastry and Amanda Askell and Pamela Mishkin and Jack Clark and Gretchen Krueger and Ilya Sutskever},
      year={2021},
      eprint={2103.00020},
      archivePrefix={arXiv},
      primaryClass={cs.CV},
      url={https://arxiv.org/abs/2103.00020}, 
}

@misc{faysse2025colpaliefficientdocumentretrieval,
      title={ColPali: Efficient Document Retrieval with Vision Language Models}, 
      author={Manuel Faysse and Hugues Sibille and Tony Wu and Bilel Omrani and Gautier Viaud and Céline Hudelot and Pierre Colombo},
      year={2025},
      eprint={2407.01449},
      archivePrefix={arXiv},
      primaryClass={cs.IR},
      url={https://arxiv.org/abs/2407.01449}, 
}
\clearpage



\section{Appendix}

In this subsection, we present lucid examples of questions across various tasks of our benchmark \benchmarkname.

\begin{figure*}[h]
    \centering
    \includegraphics[width=0.95\textwidth]{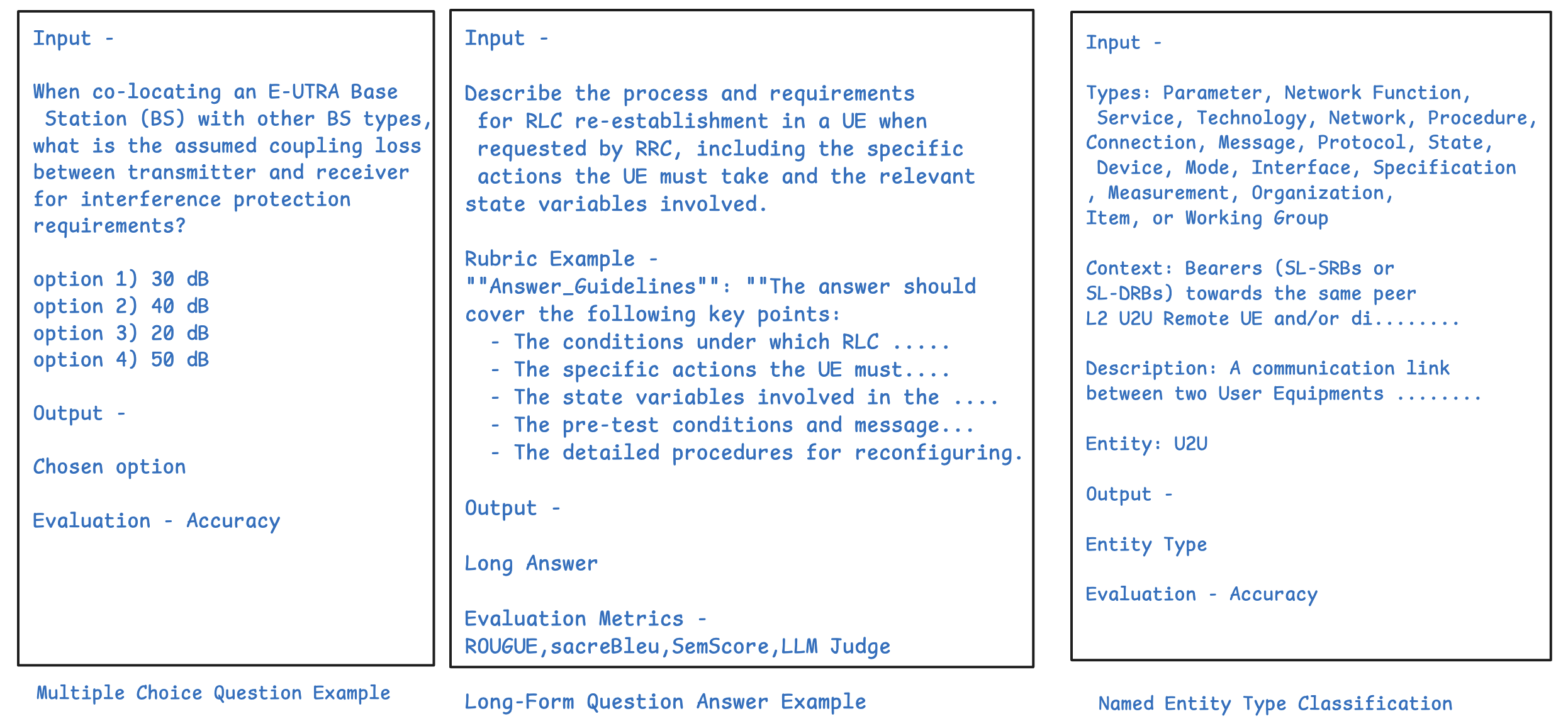}
    \caption{Examples of Multiple Choice Questions, Long Answer Questions, and Named Entity Classification questions from the text category.}
    \label{fig:text_example}
\end{figure*}

\begin{figure*}[h]
    \centering
    \includegraphics[width=0.95\textwidth]{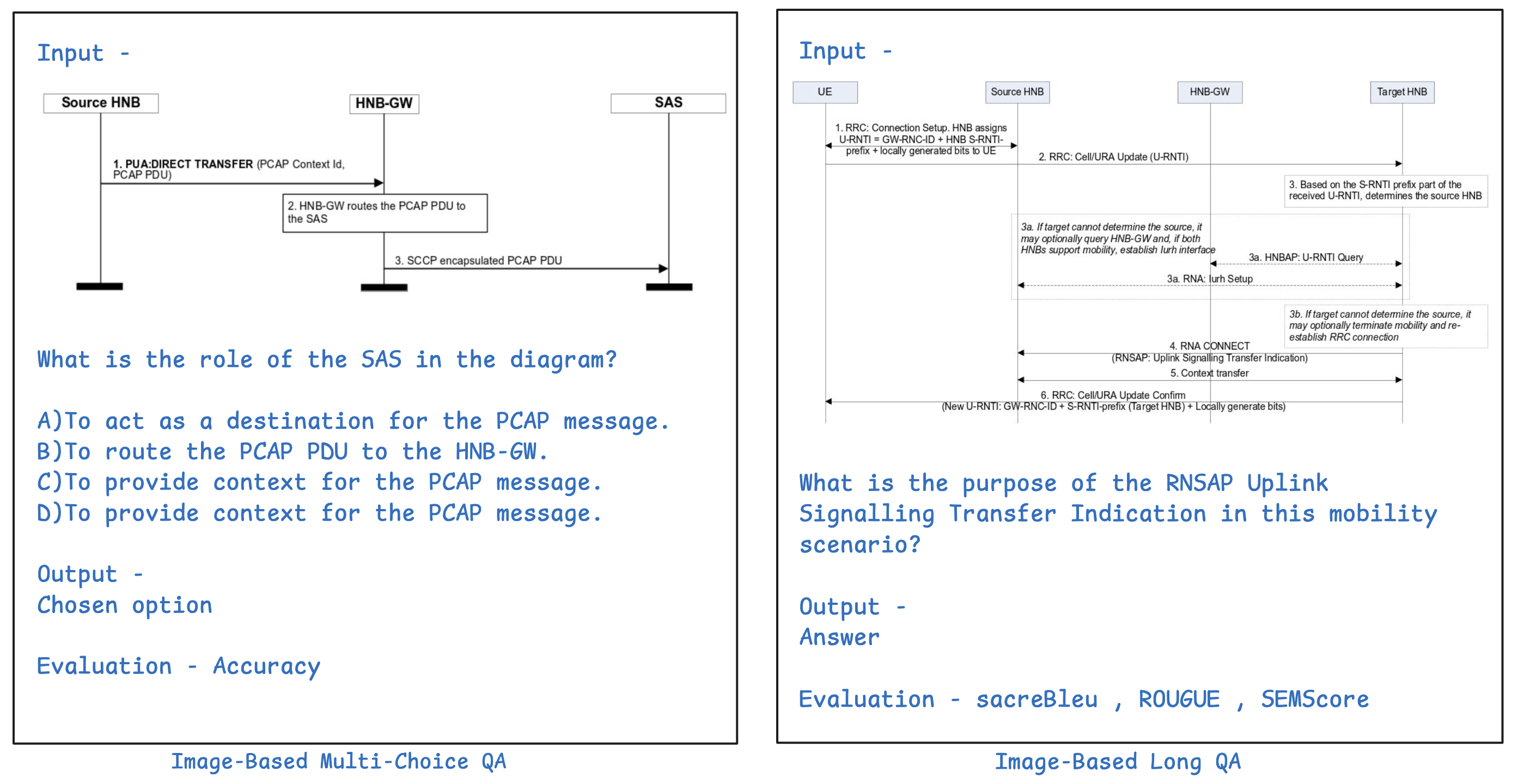}
    \caption{Examples of Image-based Multiple Choice Questions and Long Answer Questions from the image category.}
    \label{fig:image_example}
\end{figure*}

\begin{figure}[h]
    \centering
    \includegraphics[scale=0.23]{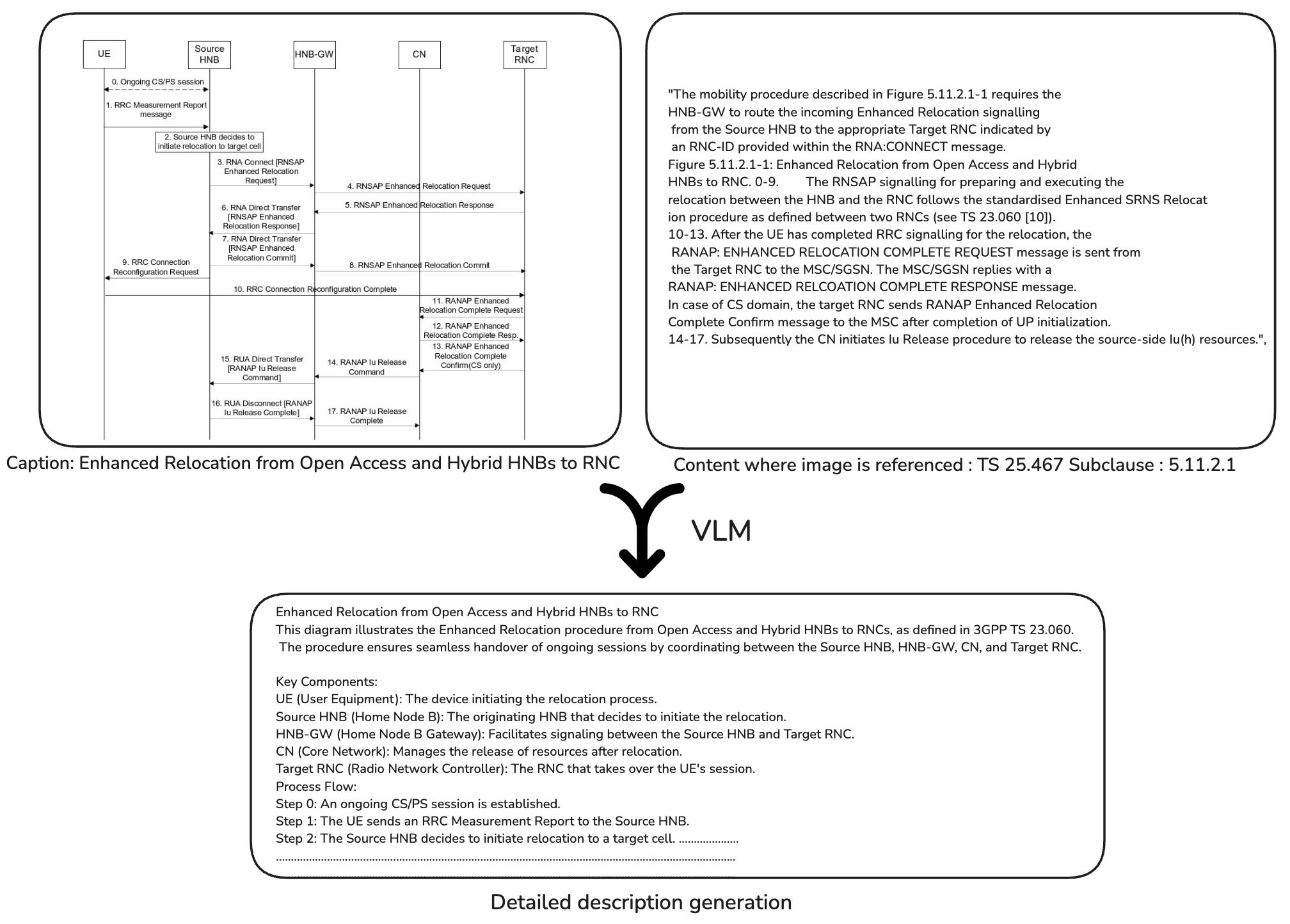}
    \caption{Example of the process used to generate detailed image descriptions for image-based tasks.}
    \label{fig:process}
\end{figure}

\begin{figure*}[h]
    \centering
    \includegraphics[width=0.6\textwidth]{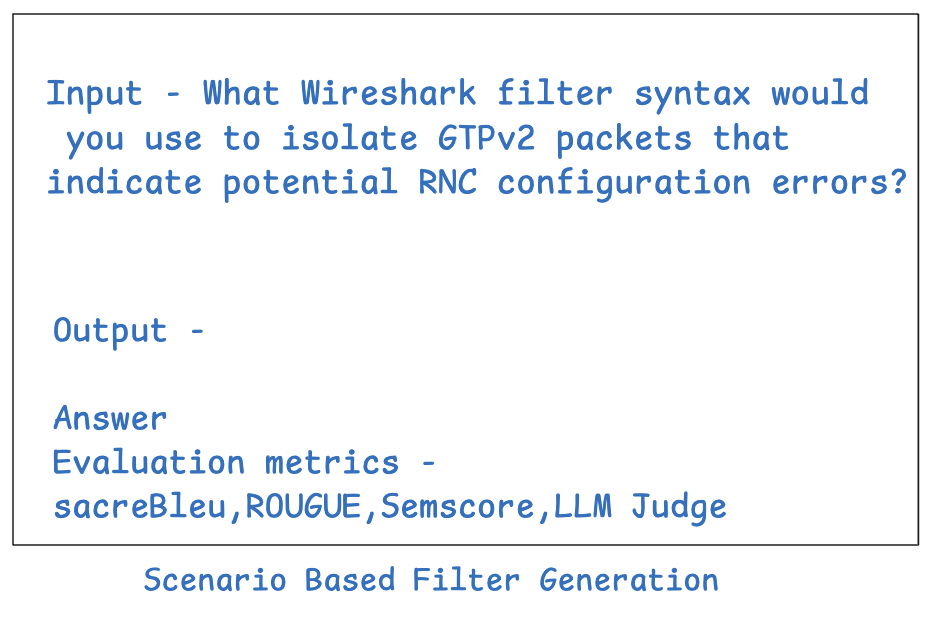}
    \caption{Example of Scenario-Based Filter Generation Benchmark.}
    \label{fig:pcap_example}
\end{figure*}

\begin{figure*}[h]
    \centering
    \includegraphics[width=0.95\textwidth]{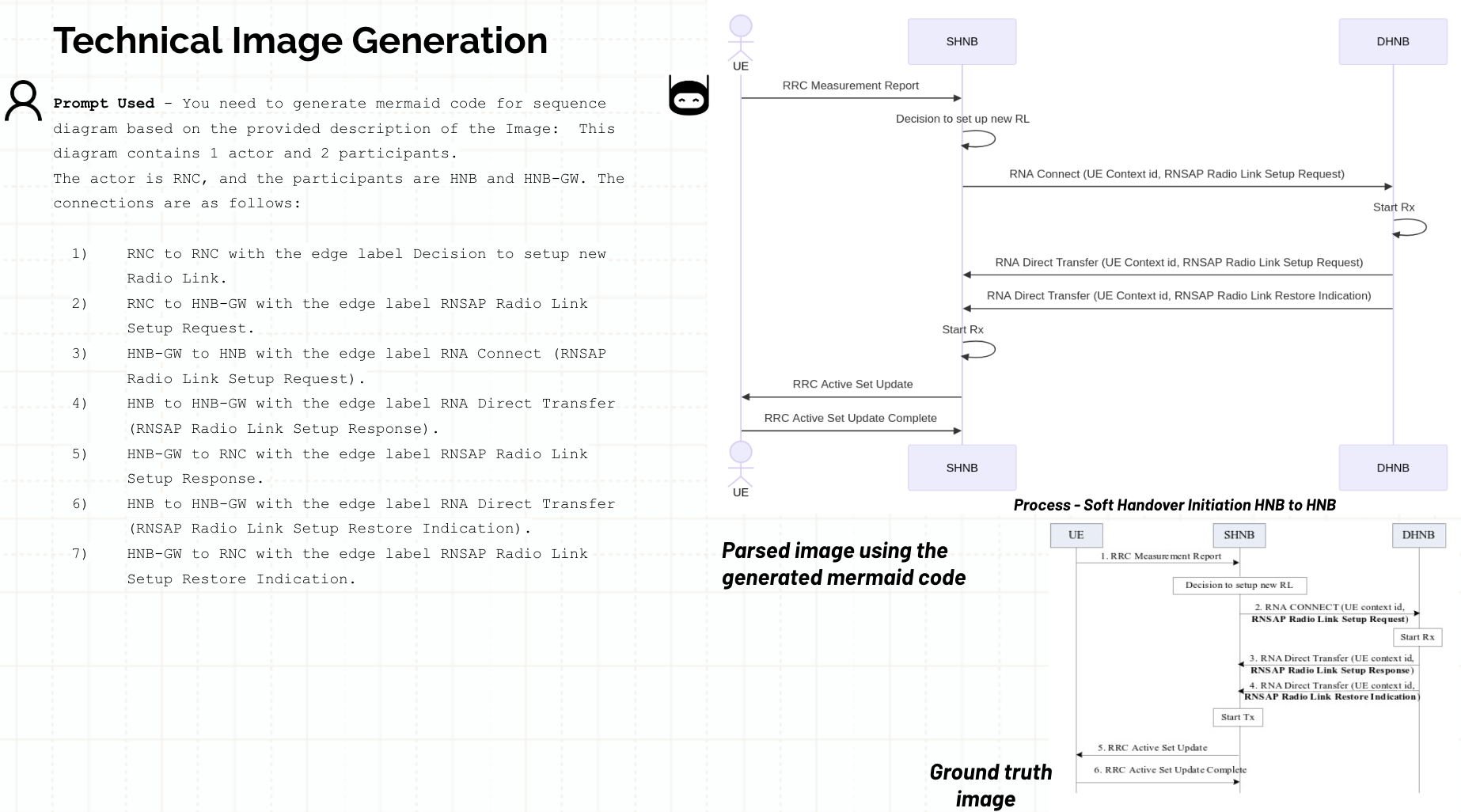}
    \caption{Illustration of how the model generates a sequence diagram from a textual prompt using Mermaid code.}
    \label{fig:image_gen_sequence}
\end{figure*}

\begin{figure*}[h]
    \centering
    \includegraphics[width=0.6\textwidth]{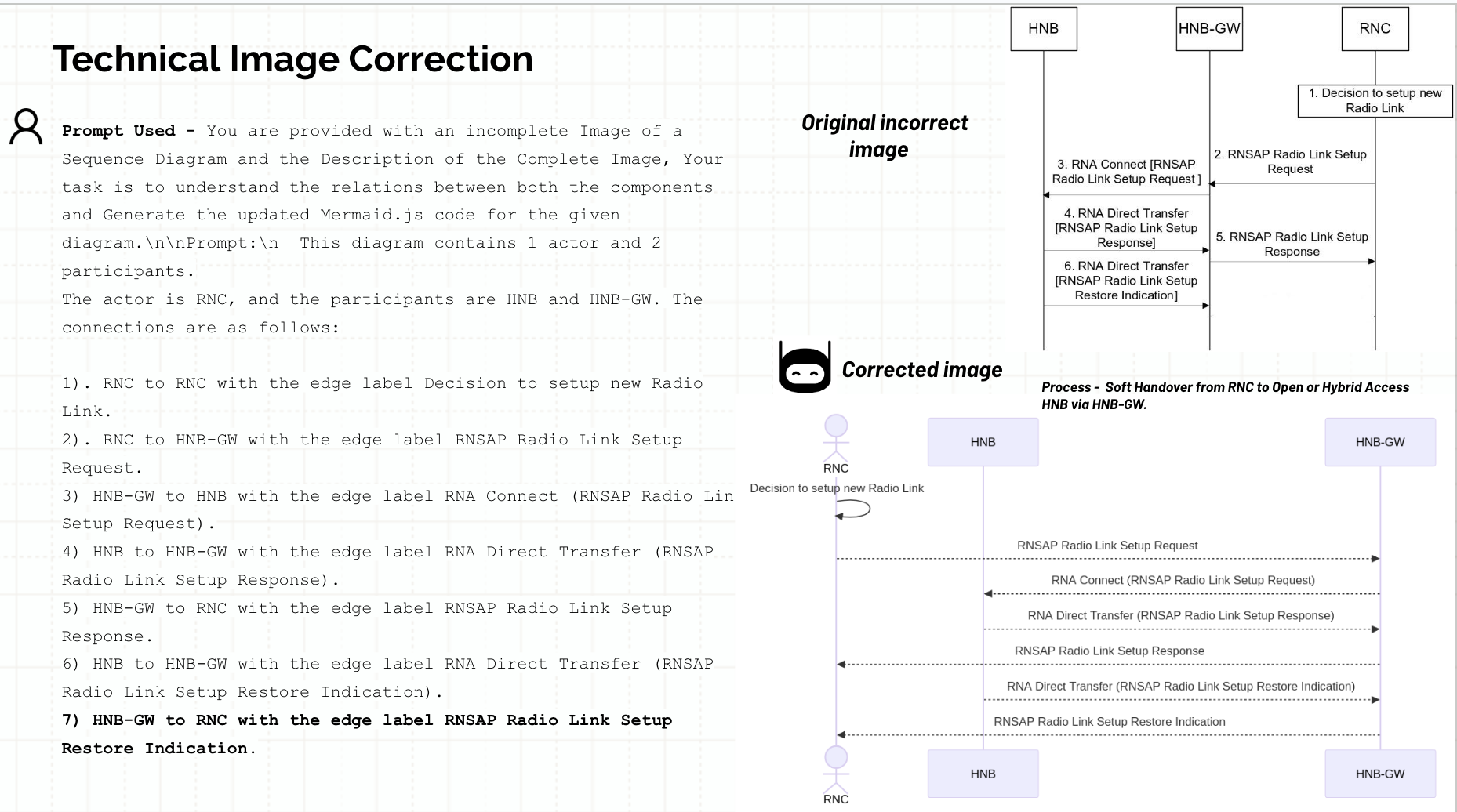}
    \caption{Model's capability to correct inaccurate sequence diagrams using Mermaid code.}
    \label{fig:image_correct_sequence}
\end{figure*}

\begin{figure*}[h]
    \centering
    \includegraphics[width=0.6\textwidth]{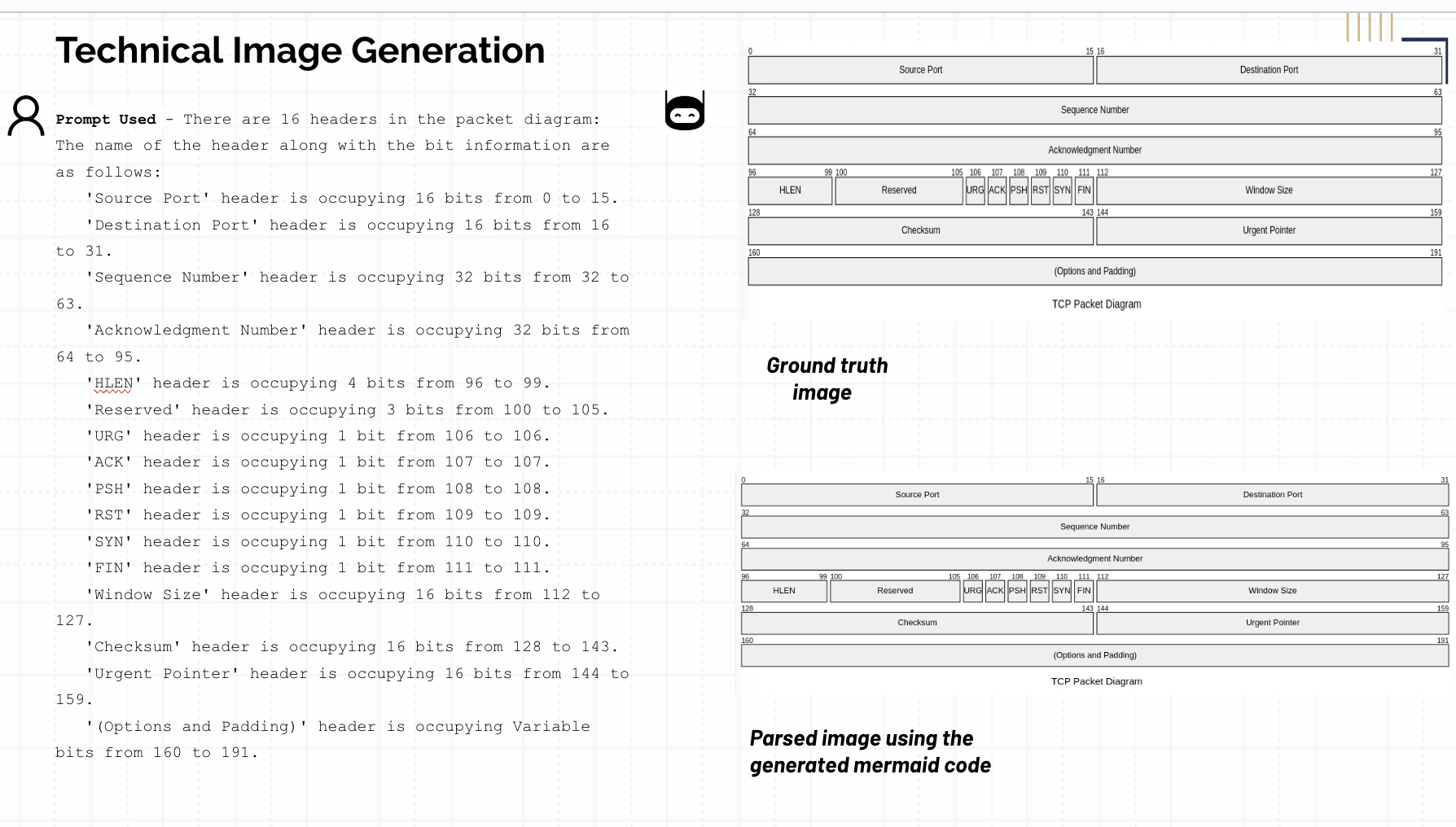}
    \caption{Model generating accurate packet diagrams from bit-level header specifications using Mermaid code.}
    \label{fig:image_gen_packet}
\end{figure*}

\begin{figure*}[h]
    \centering
    \includegraphics[width=0.6\textwidth]{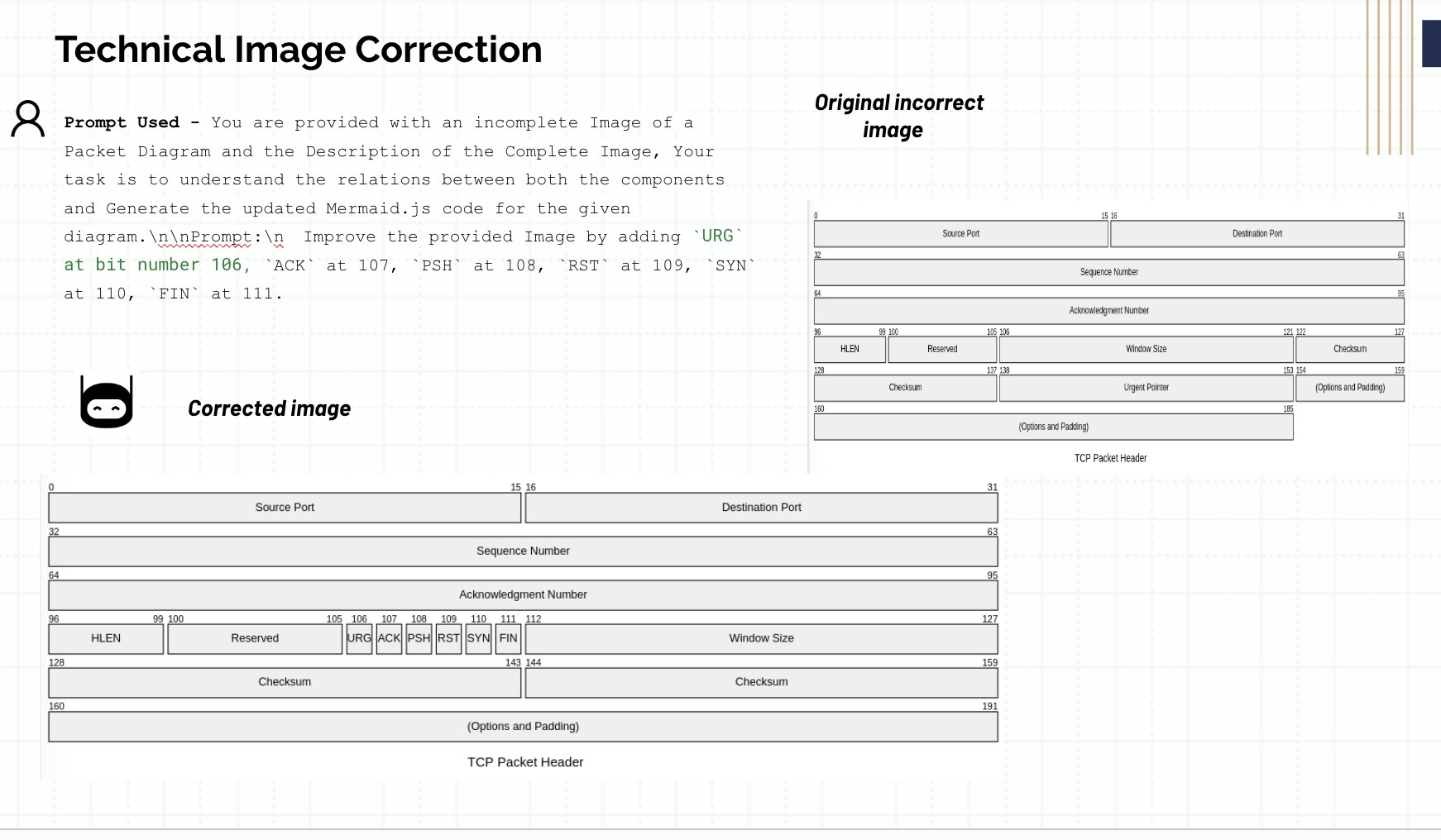}
    \caption{Correction of an incomplete TCP packet header diagram using Mermaid code.}
    \label{fig:image_corr_packet}
\end{figure*}

\begin{figure*}[h]
    \centering
    \includegraphics[width=\textwidth]{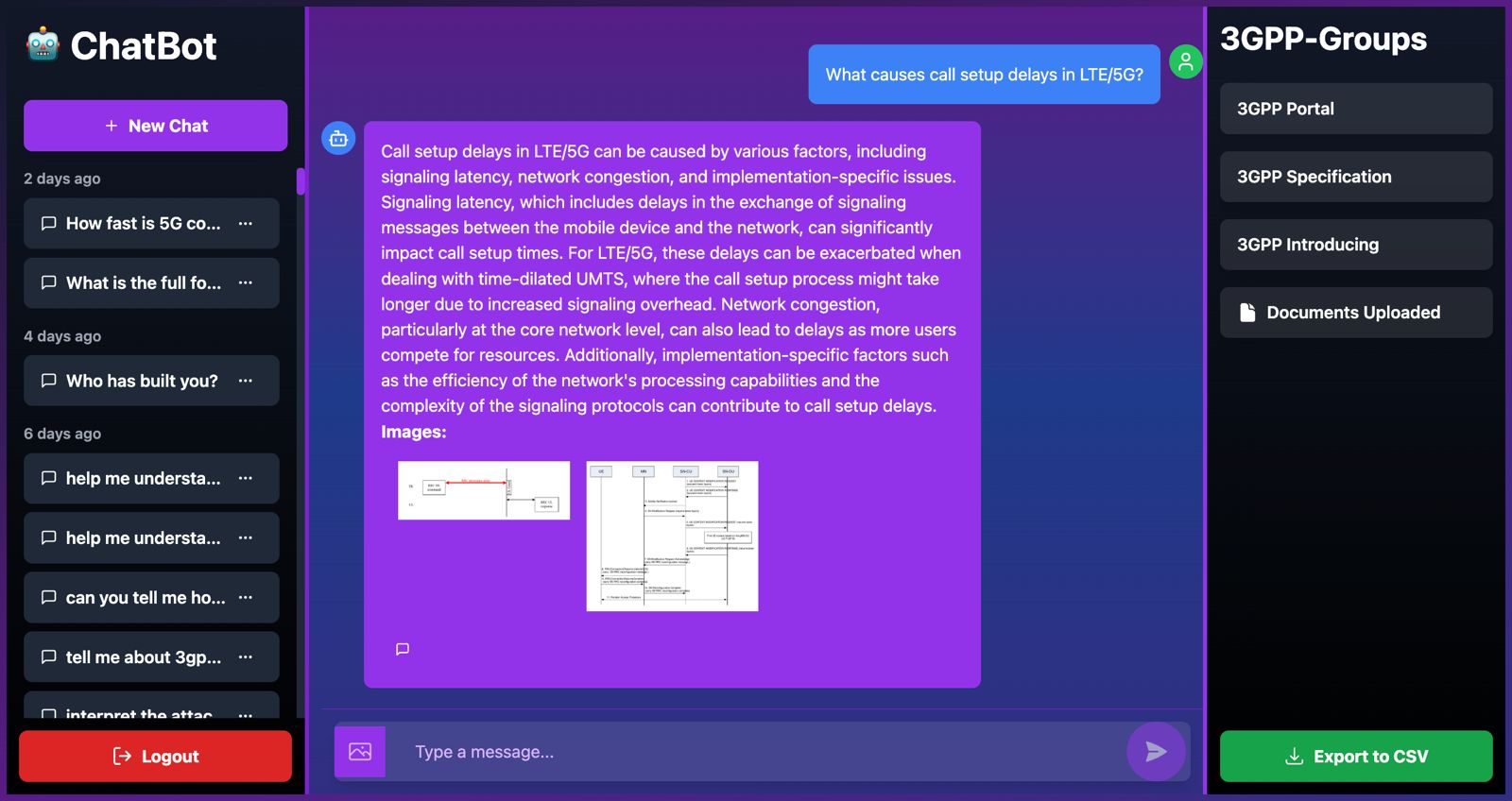}
    \caption{Example of a Multi-Modal chatbot designed for 3GPP documents.}
    \label{fig:chatbot_multimodal}
\end{figure*}

In this subsection, we present the details of the various working groups of 3GPP. Grouping results by working groups allows us to analyze the weak areas of the model.

\begin{table*}[h]
    \centering
    \begin{tabular}{|l|l|}
        \hline
        \textbf{Working Group Name} & \textbf{Domain} \\ \hline
        \textbf{TSG RAN} & Radio Access Network \\ \hline
        RAN WG1 & Radio Layer 1 (Physical layer) \\
        RAN WG2 & Radio Layer 2 and Radio Layer 3 Radio Resource Control \\
        RAN WG3 & UTRAN/E-UTRAN/NG-RAN architecture and related interfaces \\
        RAN WG4 & Radio Performance and Protocol Aspects \\
        RAN WG5 & Mobile terminal conformance testing \\
        RAN AH1 & ITU-R Ad Hoc \\ \hline
        \textbf{TSG SA} & Service \& System Aspects \\ \hline
        SA WG1 & Services \\
        SA WG2 & System Architecture and Services \\
        SA WG3 & Security and Privacy \\
        SA WG4 & Multimedia Codecs, Systems, and Services \\
        SA WG5 & Management, Orchestration, and Charging \\
        SA WG6 & Application Enablement and Critical Communication Applications \\ \hline
        \textbf{TSG CT} & Core Network \& Terminals \\ \hline
        CT WG1 & User Equipment to Core Network Protocols \\
        CT WG3 & Interworking with External Networks \& Policy and Charging Control \\
        CT WG4 & Core Network Protocols \\
        CT WG6 & Smart Card Application Aspects \\ \hline
    \end{tabular}
    \caption{3GPP Working Groups and their Domains.}
    \label{tab:3gpp_wgs}
\end{table*}


         

\end{document}